\documentclass[10pt,twocolumn,letterpaper]{article}

\usepackage{cvpr}              

\usepackage{graphicx}
\usepackage{amsmath}
\usepackage{amssymb}
\usepackage{booktabs}
\usepackage{makecell}
\usepackage{multirow}
\usepackage{comment}

\newcommand{\parsection}[1]{\noindent\textbf{#1} }

\usepackage{algorithm}
\usepackage{algpseudocode}

\algrenewcommand{\algorithmicrequire}{\textbf{Input:}}
\algrenewcommand{\algorithmicensure}{\textbf{Output:}}

\usepackage{bbding}
\usepackage{color}
\usepackage{pifont}
\usepackage{xcolor}
\usepackage{colortbl}
\newcommand{\cmark}{\ding{51}\xspace}%
\newcommand{\cmarkg}{\textcolor{gray}{\ding{51}}\xspace}%
\newcommand{\xmark}{\ding{55}\xspace}%
\newcommand{\xmarkr}{\textcolor{red}{\ding{55}}\xspace}%

\newcommand{\modelname}{MaskFreeVIS\xspace} 
\newcommand{\lossname}{Temporal KNN-patch Loss\xspace} 
\usepackage[pagebackref,breaklinks,colorlinks]{hyperref}

\usepackage[capitalize]{cleveref}
\crefname{section}{Sec.}{Secs.}
\Crefname{section}{Section}{Sections}
\Crefname{table}{Table}{Tables}
\crefname{table}{Tab.}{Tabs.}

\def\cvprPaperID{706} 
\def\confName{CVPR}
\def\confYear{2023}

\begin{document}

\title{Mask-Free Video Instance Segmentation}
\author{
 Lei Ke$^{1,2}$\hspace{0.35cm}Martin Danelljan$^1$\hspace{0.35cm}Henghui Ding$^1$\hspace{0.35cm}Yu-Wing Tai$^2$\hspace{0.35cm}Chi-Keung Tang$^2$\hspace{0.35cm}Fisher Yu$^1$\\
 $^1$ETH Z{\"u}rich\hspace{1.5cm}$^2$HKUST\hspace{1.5cm} \\
 }
 
\twocolumn[{%
\renewcommand\twocolumn[1][]{#1}%
\maketitle 
\begin{center} 
\vspace{-0.1in}
\centering 
\includegraphics[width=0.996\textwidth]{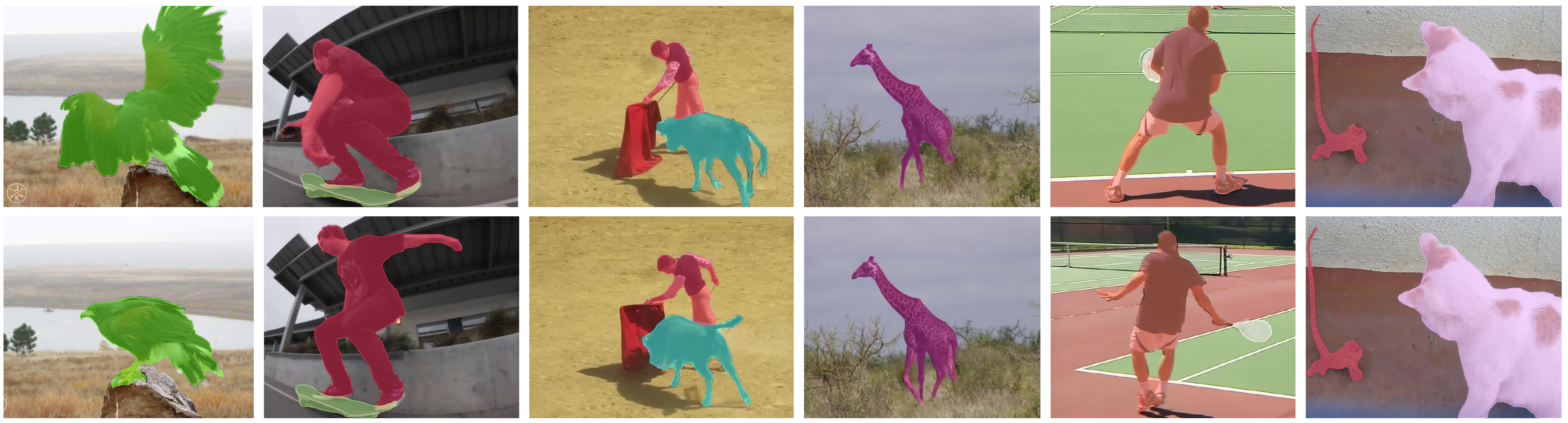}
\vspace{-6mm}
\captionof{figure}{Video instance segmentation (VIS) results of our \modelname, trained \textbf{without} using any video or image mask annotation. By achieving a remarkable 42.5\% mask AP on the YouTube-VIS \textit{val} dataset, with a ResNet-50 backbone, our approach demonstrates that \textbf{high-performing VIS can be learned even without any mask annotations}.} 
\label{fig:teaser}
\vspace{2mm}
\end{center}
}]

\begin{abstract}
  The recent advancement in Video Instance Segmentation (VIS) has largely been driven by the use of deeper and increasingly data-hungry transformer-based models.
  However, video masks are tedious and expensive to annotate, limiting the scale and diversity of existing VIS datasets.
  In this work, we aim to remove the mask-annotation requirement.
  We propose MaskFreeVIS, achieving highly competitive VIS performance, while only using bounding box annotations for the object state.
  We leverage the rich temporal mask consistency constraints in videos by introducing the Temporal KNN-patch Loss (TK-Loss), providing strong mask supervision without any labels. 
  Our TK-Loss finds one-to-many matches across frames, through an efficient patch-matching step followed by a K-nearest neighbor selection. 
  A consistency loss is then enforced on the found matches. 
  Our mask-free objective is simple to implement, has no trainable parameters, is computationally efficient, yet outperforms baselines employing, \eg, state-of-the-art optical flow to enforce temporal mask consistency. 
  We validate MaskFreeVIS on the YouTube-VIS 2019/2021, OVIS and BDD100K MOTS benchmarks. The results clearly demonstrate the efficacy of our method by drastically narrowing the gap between fully and weakly-supervised VIS performance.
  Our code and trained models are available at \url{https://github.com/SysCV/MaskFreeVis}.

\end{abstract}


\section{Introduction}
\label{sec:intro}



Video Instance Segmentation (VIS) requires jointly detecting, tracking and segmenting all objects in a video from a given set of categories.
To perform this challenging task, state-of-the-art VIS models are trained with complete video annotations from VIS datasets~\cite{yang2019video,qi2022occluded,bdd100k}.
However, video annotation is costly, in particular regarding object mask labels. Even coarse polygon-based mask annotation is multiple times slower than annotating video bounding boxes~\cite{cheng2022pointly}. 
Expensive mask annotation makes existing VIS benchmarks difficult to scale, limiting the number of object categories covered. 
This is particularly a problem for the recent transformer-based VIS models~\cite{cheng2021mask2former,wu2021seqformer,vmt}, which tend to be exceptionally data-hungry.
We therefore revisit the need for complete mask annotation by studying the problem of weakly supervised VIS \textit{under the mask-free setting}.

While there exist box-supervised instance segmentation models~\cite{tian2021boxinst,hsu2019weakly,lan2021discobox,li2022box}, they are designed for images. These weakly-supervised single-image methods do not utilize temporal cues when learning mask prediction, leading to lower accuracy when directly applied to videos.
As a source for weakly supervised learning, videos contain much richer information about the scene. In particular, videos adhere to the temporal mask consistency constraint, where the regions corresponding to the same underlying object across different frames should have the same mask label. In this work, we set out to leverage this important constraint for mask-free learning of VIS.

\begin{figure}[!t]
	\centering
	\vspace{-0.1in}
	\includegraphics[width=1.0\linewidth]{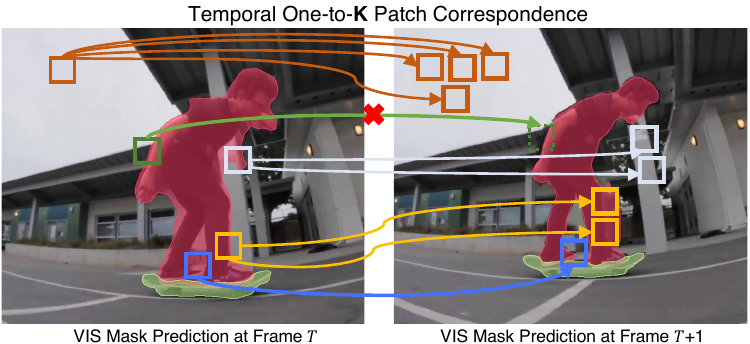}
	\vspace{-0.25in}
	\caption{Our \lossname enforces mask consistency between  one-to-$k$ patch correspondences found across frames, which allow us to cover the cases where: \textbf{(i)} A unique one-to-one match exists (blue); \textbf{(ii)} Multiple matches are found due to ambiguities in homogenuous regions (orange) or along image edges (white and yellow); \textbf{(iii)} No match is found due to \eg occlusions (green). This allows us to robustly leverage mask consistency constraints in challenging videos.}
	\label{fig:teaser_exp}
	\vspace{-0.25in}
\end{figure}


We propose \modelname method, for high performance VIS without any mask annotations. 
To leverage temporal mask consistency, we introduce the~\lossname (TK-Loss), as in Figure~\ref{fig:teaser_exp}. 
To find regions corresponding to the same underlying video object, our TK-Loss first builds correspondences across frames by patch-wise matching.
For each target patch, only the top $K$ matches in the neighboring frame with high enough matching score are selected. 
A temporal consistency loss is then applied to all found matches to promote the mask consistency.
Specifically, our surrogate objective function not only promotes the one-to-$k$ matched regions to reach the same mask probabilities, but also commits their mask prediction to a confident foreground or background prediction by entropy minimization. 
Unlike flow-based models~\cite{teed2020raft,liu2021weakly}, which assume one-to-one matching, our approach builds robust and flexible one-to-$k$ correspondences to cope with \eg occlusions and homogeneous regions, \textit{without} introducing additional model parameters or inference cost.

The TK-Loss is easily integrated into existing VIS methods, with no architecture modifications required. During training, our TK-Loss simply replaces the conventional video mask losses in supervising video mask generation. To further enforce temporal consistency through the video clip, TK-Loss is employed in a cyclic manner instead of using dense frame-wise connections. This greatly reduces memory cost with negligible performance drop.

We extensively evaluate our \modelname on four large-scale VIS benchmarks, \ie, YouTube-VIS 2019/2021~\cite{yang2019video}, OVIS~\cite{qi2022occluded}, and BDD100K MOTS~\cite{bdd100k}. \modelname achieves competitive VIS performance \textbf{without} using any video masks or even image mask labels on all datasets. Validated on various methods and backbones, \modelname achieves 91.25\% performance of its fully supervised counterparts, even outperforming a few recent fully-supervised methods~\cite{wu2022efficient,inspro,hwang2021video,pcan} on the popular YTVIS benchmark.
Our simple yet effective design greatly narrows the performance gap between weakly-supervised and fully-supervised video instance segmentation. It further demonstrates that expensive video masks, or even image masks, is not necessary for training high-performing VIS models.  

\begin{table}[!t]
\footnotesize
	\caption{Mask annotation requirement for state-of-the-art VIS methods. Results are reported using ResNet-50 as backbone on the YTVIS 2019~\cite{yang2019video} benchmark. \textbf{Video Mask}: using YTVIS video mask labels. \textbf{Image Mask}: using COCO~\cite{lin2014microsoft} image mask labels for image-based pretraining. \textbf{Pseudo Video}: using Pseudo Videos from COCO images for joint training~\cite{wu2021seqformer}. \modelname achieves \underline{91.5\% (42.5 \vs 46.4)} of its fully-supervised baseline performance (Mask2Former) \textbf{without} using any masks during training.}
	\vspace{-0.1in}
	\centering
        \setlength\tabcolsep{8.5pt}
		\begin{tabular}{lcccc}
					\toprule
					Method & \makecell[c]{\textbf{Video}\\ \textbf{Mask}} & \makecell[c]{\textbf{Image}\\ \textbf{Mask}} & \makecell[c]{\textbf{Pseudo}\\ \textbf{Video}} & AP \\
					\midrule
					SeqFormer~\cite{wu2021seqformer} & \cmarkg & \cmarkg & \cmarkg & 47.4 \\
					VMT~\cite{vmt} & \cmarkg & \cmarkg & \cmarkg & 47.9 \\
					\midrule
			  	   Mask2Former~\cite{cheng2021mask2former} & \cmarkg & \cmarkg & \cmarkg  & 47.8 \\
					\rowcolor{lightgray!16}\textbf{\modelname (ours)} & \xmarkr & \cmarkg & \cmarkg & 46.6 \\ 
					\midrule
			  	   Mask2Former~\cite{cheng2021mask2former} & \cmarkg & \cmarkg & \xmark  & 46.4 \\
					\rowcolor{magenta!16}\textbf{\modelname (ours)} & \xmarkr & \xmarkr & \xmark & 42.5 \\ 
					\bottomrule
			\end{tabular}
	\vspace{-0.2in}
	\label{tab:teaser_table}
\end{table}

Our contributions are summarized as follows: 
\textbf{(i)} To utilize temporal information, we develop a new parameter-free \lossname, which leverages temporal masks consistency using unsupervised one-to-$k$ patch correspondence. We extensively analyze the TK-Loss through ablative experiments.
\textbf{(ii)} Based on the TK-Loss, we develop the \modelname method, enabling training existing state-of-the-art VIS models \textit{without} any mask annotation. 
\textbf{(iii)} To the best of our knowledge, \modelname is the first mask-free VIS method attaining high-performing segmentation results. 
We provide qualitative results in Figure~\ref{fig:teaser}.
As in Table~\ref{tab:teaser_table}, when integrated into the Mask2Former~\cite{cheng2021mask2former} baseline with ResNet-50, our \modelname achieves 42.5\% mask AP on the challenging YTVIS 2019 benchmark while using \textbf{no} video or image mask annotations. Our approach further scales to larger backbones, achieving 55.3\% mask AP on Swin-L backbone with~\textit{no} video mask annotations.

We hope our approach will facilitate achieving label-efficient video instance segmentation, enabling building even larger-scale VIS benchmarks with diverse categories by lifting the mask annotation restriction.

\section{Related Work}
\parsection{Video Instance Segmentation (VIS)} Existing VIS methods can be summarized into three categories: two-stage, one-stage, and transformer-based. Two-stage approaches~\cite{yang2019video,bertasius2020classifying,lin2020video,pcan,lin2021video} extend the Mask R-CNN family~\cite{he2017mask,ke2021bcnet} by designing an additional tracking branch for object association. One-stage works~\cite{CaoSipMask_ECCV_2020,STMask-CVPR2021,liu2021sg,Yang_2021_ICCV} adopt anchor-free detectors~\cite{tian2019fcos}, generally using linear masks basis combination~\cite{yolact-iccv2019} or conditional mask prediction generation~\cite{tian2020conditional}. 
For the transformer-based models~\cite{cheng2021mask2former,wu2021seqformer,yang2022tevit,mssts2022,heo2022vita}, VisTr~\cite{wang2020end} firstly adapts the transformer~\cite{carion2020end} for VIS, and IFC~\cite{hwang2021video} further improves its efficiency via memory tokens.
Seqformer~\cite{wu2021seqformer} proposes frame query decomposition while Mask2Former~\cite{cheng2021mask2former} includes  masked attention. VMT~\cite{vmt} extends Mask Transfiner~\cite{transfiner} to video for high-quality VIS, and IDOL~\cite{IDOL} focuses on online VIS. State-of-the-art VIS methods with growing capacity put limited emphasis on weak supervision. In contrast, the proposed~\modelname is the first method targeting mask-free VIS while attaining competitive performance.

\parsection{Multiple Object Tracking and Segmentation (MOTS)} Most MOTS methods~\cite{voigtlaender2019mots,milan2015joint,meinhardt2022trackformer, Athar_Mahadevan20ECCV,wu2021track} follow the tracking-by-detection principle. PCAN~\cite{pcan} improves temporal segmentation by utilizing space-time memory prototypes, while the one-stage method Unicorn~\cite{unicorn} focuses on unification of different tracking frameworks. Compared to the aforementioned fully-supervised MOTS methods, \modelname focuses on label efficient training without GT masks by proposing a new surrogate temporal loss which can be easily integrated on them.

\parsection{Mask-Free VIS}
Most mask-free instance segmentation works~\cite{lee2021bbam,khoreva2017simple,hsu2019weakly,li2022box,papandreou2015weakly,rajchl2016deepcut,song2019box,cheng2022pointly} are designed for single images and thus neglect temporal information.
Earlier works BoxSup~\cite{dai2015boxsup} and Box2Seg~\cite{kulharia2020box2seg} rely on region proposals produced by MCG~\cite{pont2016multiscale} or GrabCut~\cite{rother2004grabcut}, leading to slow training.
BoxInst~\cite{tian2021boxinst} proposes the surrogate projection and pixel pairwise losses to replace the original mask learning loss of CondInst~\cite{tian2020conditional}, while DiscoBox~\cite{lan2021discobox} focuses on generating pseudo mask labels guided by a teacher model. 

Earlier works have investigated the use of videos for weakly-, semi-, or un-supervised segmentation by leveraging motion or temporal consistency \cite{tokmakov2016learning,tsai2016semantic,kipf2022conditional}. 
Most aforementioned approaches do not address the VIS problem, and use optical flow for frame-to-frame matching~\cite{liu2021weakly,lee2019frame,saleh2017bringing}. In particular, FlowIRN~\cite{liu2021weakly} explores VIS using only classification labels and incorporates optical flow to leverage mask consistency. 
The limited performance makes the class-label only or fully-unsupervised setting difficult to deploy in the real world.
SOLO-Track~\cite{fu2021learning} aims to train VIS models without video annotations, and one concurrent work MinVIS~\cite{huang2022minvis} performs VIS without video-based model architectures.
Unlike the above weakly-supervised training settings, our \modelname is designed for eliminating the mask annotation requirement for VIS, as we note that video mask labeling is particularly expensive.
\modelname enables training VIS models {\em without} any video masks, or even image masks. 
Despite its simplicity,~\modelname drastically reduces the gap between fully-supervised and weakly-supervised VIS models, making weakly-supervised models more accessible in practice.

\section{Method}
We propose \modelname to tackle video instance segmentation (VIS) \textbf{without} using any video or even image mask labels. Our approach is generic and can be directly applied to train state-of-the-art VIS methods, such as Mask2Former~\cite{cheng2021mask2former} and SeqFormer~\cite{wu2021seqformer}. 
In \cref{sec:method}, we first present the core component of \modelname: the \lossname (TK-Loss), which leverages temporal consistency to supervise accurate mask prediction, without any human mask annotations.
In \cref{sec:training}, we then describe how to integrate the TK-Loss with existing spatial weak segmentation losses for transformer-based VIS methods, to achieve mask-free training of VIS approaches.
Finally, we introduce image-based pretraining details of our \modelname in \cref{sec:imagetraining}.

\subsection{\modelname}
\label{sec:method}
\begin{figure*}[!t]
	\centering
	\vspace{-0.2in}
	\includegraphics[width=1.0\linewidth]{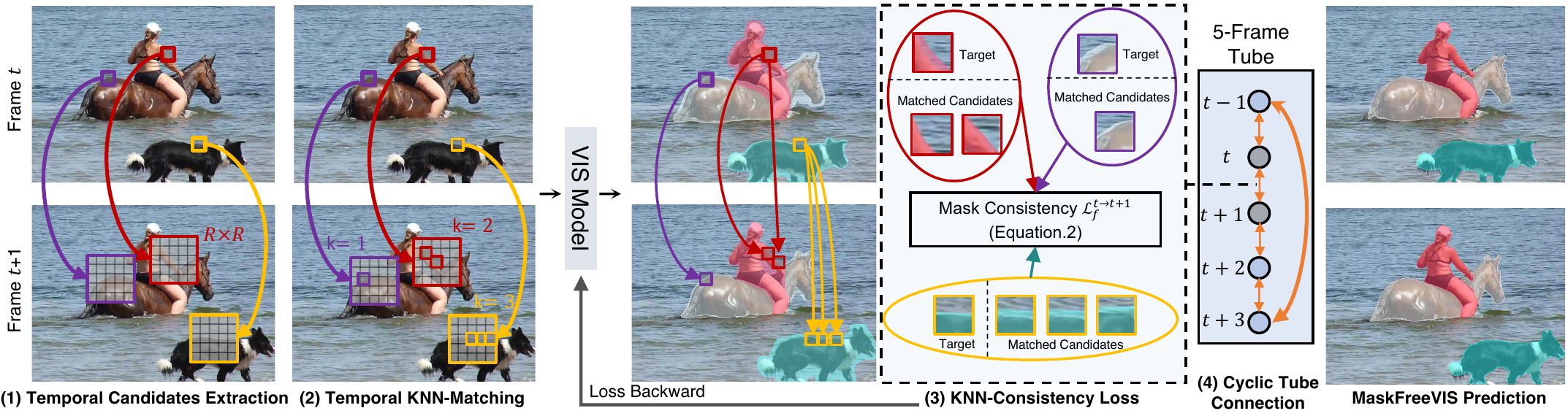}
	\vspace{-0.28in}
	\caption{\lossname has four steps: \textbf{1)} Patch Candidate Extraction: Patch candidates searching across frames with radius $R$. \textbf{2)} Temporal KNN-Matching: Match $k$ high-confidence candidates by patch affinities. \textbf{3)} Consistency loss: Enforce mask consistency objective (Eq.~\ref{eq:eq2}) among the matches. \textbf{4)} Cyclic Tube Connection: Temporal loss aggregation in the 5-frame tube, detailed in Figure~\ref{fig:tube_connect}.}
	\label{fig:tkloss}
	\vspace{-0.2in}
\end{figure*}

In this section, we introduce the \lossname, illustrated in Figure~\ref{fig:tkloss}. 
It serves as an unsupervised objective for mask prediction that leverages the rich spatio-temporal consistency constraints in unlabelled videos.

\subsubsection{Temporal Mask Consistency}
While an image constitutes a single snapshot of a scene, a video provides multiple snapshots displaced in time.
Thereby, a video depicts continuous \emph{change} in the scene. 
Objects and background move, deform, are occluded, experience variations in lighting, motion blur, and noise, leading to a sequence of different images that are closely related through gradual transformations.

Consider a small region in the scene (\cref{fig:teaser_exp}), belonging either to an object or background. The pixels corresponding to the projection of this region should have the same mask prediction in every frame, as they belong to the same underlying physical object or background region. However, the aforementioned dynamic changes in the video lead to substantial appearance variations, serving as a natural form of data augmentation. The fact that the pixels corresponding to the same underlying object region should have the same mask prediction under temporal change therefore provides a powerful constraint, \ie, \emph{temporal mask consistency}, which can be used for mask supervision~\cite{liu2021weakly,lee2019frame,tokmakov2016learning,tsai2016semantic,kipf2022conditional}.

The difficulty in leveraging the temporal mask consistency constraint stems from the problem of establishing reliable correspondences between video frames. 
Objects often undergo fast motion, deformations, etc., resulting in substantial appearance change. 
Furthermore, regions in the scene may be occluded or move out of the image from one frame to the other. In such cases, no correspondence exist. Lastly, videos are often dominated by homogenous regions, such as sky and ground, where the establishment of one-to-one correspondences are error-prone or even ill-defined. 

The problem of establishing dense one-to-one correspondences between subsequent video frames, known as optical flow, is a long-standing and popular research topic. However, when attempting to enforce temporal mask consistency through optical flow estimation~\cite{liu2021weakly,lee2019frame,saleh2017bringing}, one encounters two key problems.
1) The one-to-one assumption of optical flow is not suitable in cases of occlusions, homogenous regions, and along single edges, where the correspondence is either nonexistent, undefined, ambiguous, uncertain, or very difficult to determine. 
2) State-of-the-art optical flow estimation rely on large and complex deep networks, with large computational and memory requirements. 


Instead of using optical flow, we aim to design a simple, efficient, and parameter-free approach that effectively enforces the temporal mask consistency constraint. 
\vspace{-0.15in}

\subsubsection{Temporal KNN-patch Loss}
\label{sec:tkloss}



Our \lossname (TK-Loss) is based on a simple and flexible correspondence estimation across frames. In contrast to optical flow, we do not restrict our formulation to one-to-one correspondences. Instead, we establish one-to-$K$ correspondences. This include the conventional one-to-one ($K=1$), where a unique well-defined match exists. However, this allows us to also handle the cases of nonexistent correspondences ($K=0$) in case of occlusions, and one-to-many ($K\geq2$) in case of homogenous regions. In cases where multiple matches are found, these most often belong to the same underlying object or background due to their similar appearance, as in Figure~\ref{fig:teaser_exp}. This further benefits our mask consistency objective through denser supervision. Lastly, our approach is simple to implement, with negligible computational overhead and no learnable parameters. 
Our approach is in Figure~\ref{fig:tkloss}, and contains four main steps, which are detailed next.









\parsection{\textbf{1)} Patch Candidate Extraction:} Let $X^t_{p}$ denote an $N\times N$ target image patch centered at spatial location $p=(x, y)$ in frame $t$. Our aim is to find a set of corresponding positions $\mathcal{S}_{p}^{t\rightarrow \hat{t}}=\{\hat{p}_{i}\}_i$ in frame number $\hat{t}$ that represent the same object region.
To this end, we first select candidate locations $\hat{p}$ within a radius $R$ such that $\|p - \hat{p}\| \leq R$. 
Such windowed patch search exploits spatial proximity across neighboring frames in order to avoid an exhaustive global search.
For a fast implementation, the windowed search is performed for all target image patches $X^t_{p}$ in parallel. 

\parsection{\textbf{2)} Temporal KNN-Matching:} 
We match patch candidate patches through a simple distance computation,
\begin{equation}
\mathbf{d}^{t\rightarrow \hat{t}}_{p\rightarrow{\hat{p}}} = {\left \| X^t_{p}-X^{\hat{t}}_{\hat{p}} \right \|}, 
\end{equation}

In our ablative experiments (Sec.~\ref{sec:ablation}), we found the $L_2$ norm to be the most effective patch matching metric. 
We select the top $K$ matches with smallest patch distance $\mathbf{d}^{t\rightarrow \hat{t}}_{{p\rightarrow{\hat{p}}}}$.
Lastly low-confidence matches are removed by enforcing a maximal patch distance $D$ as $\mathbf{d}^{t\rightarrow \hat{t}}_{{p\rightarrow{\hat{p}}}} < D$. The remaining matches form the set $\mathcal{S}_{p}^{t\rightarrow \hat{t}}=\{\hat{p}_{i}\}_i$ for each location $p$.

\parsection{\textbf{3)} Consistency loss:} 
Let $M^t_p \in [0, 1]$ denote the predicted binary instance mask of an object, evaluated at position $p$ in frame $t$.
To ensure temporal mask consistency constraints, we penalize inconsistent mask predictions between a spatio-temporal point $(p,t)$ and its estimated corresponding points in $\mathcal{S}_{p}^{t\rightarrow \hat{t}}$.
In particular we use the following objective,
\begin{equation}
\label{eq:eq2}
\mathcal{L}_f^{t\to \hat{t}} = \frac{1}{HW}\sum_{p}\sum_{\hat{p}_i \in \mathcal{S}_{p}^{t\rightarrow \hat{t}}} L_\text{cons}(M_{p}^t, M_{\hat{p}_i}^{\hat{t}}),
\end{equation}
where mask consistency is measured as
\begin{equation}
\label{eq:consistent}
L_\text{cons}(M_{p}^t, M_{\hat{p}}^{\hat{t}}) = -\text{log}\big(M_{p}^t M_{\hat{p}}^{\hat{t}}+(1-M_{p}^t)(1-M_{\hat{p}}^{\hat{t}})\big) \,.
\end{equation}
Note that \cref{eq:consistent} only attains its minimum value of zero if both predictions indicate exactly background ($M_{p}^t = M_{\hat{p}}^{\hat{t}} = 0$) or foreground ($M_{p}^t = M_{\hat{p}}^{\hat{t}} = 1$). The objective does thus not only promote the two mask predictions to achieve the same probability value $M_{p}^t = M_{\hat{p}}^{\hat{t}}$, but also to commit to a certain foreground or background prediction.



\begin{figure}[!t]
	\centering
	\vspace{-0.1in}
	\includegraphics[width=1.0\linewidth]{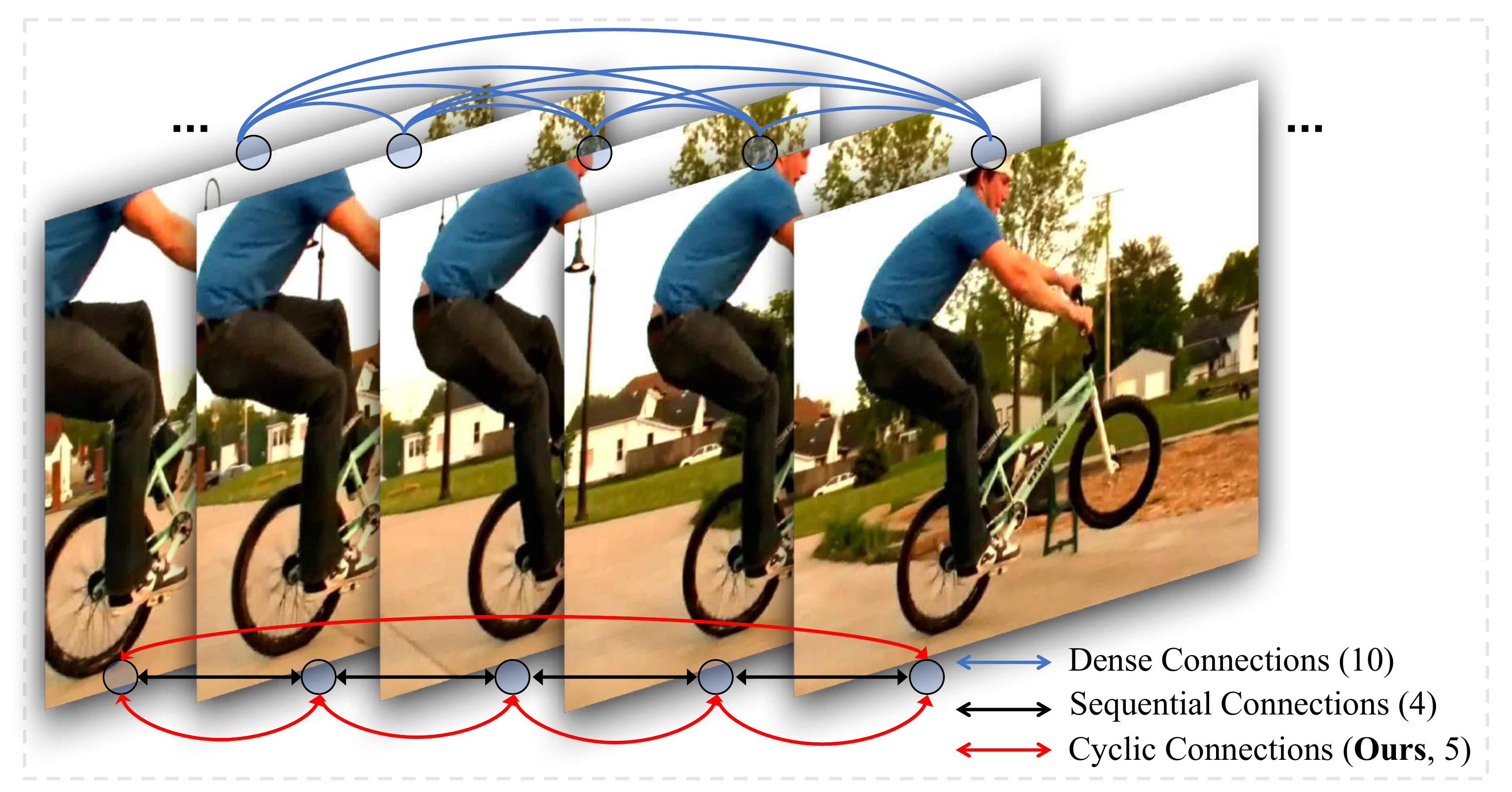}%
	\vspace{-4mm}%
	\caption{Illustration of different frame-wise tube connection settings (connection number) in the temporal loss design.}
	\label{fig:tube_connect}%
	\vspace{-3mm}
\end{figure}

\parsection{\textbf{4)} Cyclic Tube Connection:} 
Suppose the temporal tube consists of $T$ frames. We compute the temporal loss for the whole tube in a cyclic manner, as in Figure~\ref{fig:tube_connect}. The start frame is connected to the end frame, which introduces direct long-term mask consistency across the two temporally most distant frames. The temporal TK-Loss for the whole tube is given by
\begin{equation}
\small
\mathcal{L}_{\text{temp}} = 
\sum_{t=1}^{T}
\left\{ \begin{array}{ll}
 {\mathcal{L}_f^{t\to (t+1)}} &  \mbox{if  $t < T-1$} \\ 
 {\mathcal{L}_f^{t\to 0}} & \mbox{if  $t= T-1$}.
\end{array}
\right.
\end{equation}
Compared to inter-frame dense connections in Figure~\ref{fig:tube_connect}, we find the cyclic loss to achieve similar performance but greatly reduce the memory usage as validated in the experiment section.

\subsection{Training \modelname}
\label{sec:training}
In this section, we describe how to train state-of-the-art VIS methods 
using our TK-Loss,~\textbf{without} any mask annotations.
Our \modelname approach is jointly supervised with spatial-temporal surrogate losses, and is easily integrated with existing transformer-based methods. We also detail mask-free image-based pre-training for \modelname to fully eliminate mask usage during training.
\vspace{-0.1in}

\subsubsection{Joint Spatio-temporal Regularization}
To train \modelname, in addition to our proposed \lossname for temporal mask consistency, we leverage existing spatial weak segmentation losses to jointly enforce intra-frame consistency.

\parsection{Spatial Consistency}
\label{sec:spatial}
To explore spatial weak supervision signals from image bounding boxes and pixel color, we utilize the representative Box Projection Loss $L_\text{proj}$ and Pairwise Loss $L_\text{pair}$ in~\cite{tian2021boxinst}, to replace the supervised mask learning loss. 
The Projection Loss $L_\text{proj}$ enforces the projection $P'$ of the object mask onto the $\vec{x}$-axis and $\vec{y}$-axis of image to be consistent with its ground-truth box mask. For the temporal tube with $T$ frames, we concurrently optimize all predicted frame masks of the tube as,
\begin{equation}
\mathcal{L}_\text{proj} = \sum_{t=1}^{T}\sum_{d\in\{\vec{x},\vec{y}\}}D(P'_d(M^t_{p}),P'_d(M^t_{b})),
\end{equation}
where $D$ denotes dice loss, $P'$ is the projection function along $\vec{x}/\vec{y}$-axis direction, $M_p^t$ and $M_b^t$ denote predicted instance mask and its GT box mask at frame $t$ respectively. The object instance index is omitted here for clarity.

The Pairwise Loss $L_{\text{pair}}$, on the other hand, constrains spatially neighboring pixels of single frame. For pixel of locations ${p'_i}$ and ${p'_j}$ of with color similarity $\geqslant \sigma_{\text{pixel}}$, we enforce their predicted mask labels to be consistent, following \cref{eq:consistent} as,
\begin{equation}
\mathcal{L}_\text{pair} = \frac{1}{T}\sum_{t=1}^{T}\sum_{p'_i\in H\times W}L_\text{cons}(M_{p'_i}^t, M_{p'_j}^{t}).
\end{equation}
The spatial losses are combined with a weight factor $\lambda_\text{pair}$:
\begin{equation}
\mathcal{L}_{\text{spatial}} = \mathcal{L}_\text{proj} + \lambda_\text{pair} \mathcal{L}_\text{pair}.
\end{equation}

\parsection{Temporal Consistency} 
We adopt the \lossname in \cref{sec:tkloss} as $\mathcal{L}_\text{temp}$ to leverage temporal mask consistency.
The overall spatio-temporal objective $\mathcal{L}_{\text{seg}}$ for optimizing video segmentation is summarized as,
\begin{equation}
\mathcal{L}_{\text{seg}} = \mathcal{L}_{\text{spatial}} +  \lambda_\text{temp}\mathcal{L}_\text{temp}.
\end{equation}

\subsubsection{Integration with Transformer-based Methods}
\label{sec:integrate}
Existing works~\cite{hsu2019weakly,tian2020conditional} on box-supervised segmentation losses are coupled with either one-stage or two-stage detectors, such as Faster R-CNN~\cite{ren2015faster} and CondInst~\cite{tian2020conditional}, and only address the single image case.
However, state-of-the-art VIS methods~\cite{cheng2021mask2former,wu2021seqformer} are based on transformers.
These works perform object detection via set prediction, where predicted instance masks need to be matched with mask annotations when evaluating the loss.
To integrate mask-free VIS training with transformers, one key modification is in this instance-sequence matching step.

Since only ground-truth bounding boxes are available for box sequence matching, as an initial attempt, we first produce bounding box predictions from the estimated instance masks. Then, we employ the sequential box matching cost function used in VIS methods~\cite{wu2021seqformer,wang2021end}.
To compute matching cost for whole sequence, $\mathcal{L}_1$ loss and generalized IoU loss for each individual bounding box is averaged across the frames.
However, we observe the matching results of frame-wise averaging can easily be affected by a single outlier frame, especially under weak segmentation setup, leading to instability during training and performance decrease. 

\parsection{Spatio-temporal Box Mask Matching} Instead of using the aforementioned frame-wise matching, we empirically find spatio-temporal box-to-mask matching to produce substantial improvement under the weak segmentation setting. 
We first convert each predicted instance mask to a bounding box mask, and convert the ground-truth box to box mask. We then randomly sample a equal number of points from the ground-truth box mask sequence and predicted box mask sequence, respectively. Different from Mask2Former~\cite{cheng2021mask2former}, we only adopt the dice IoU loss to compute sequence matching cost. We find that cross-entropy accumulates errors per pixel, leading to imbalanced values between large and small objects. In contrast, the IoU loss in normalized per object, leading to a balanced metric. We study different instance sequence matching strategies under the mask-free VIS setting in the ablation experiments.
\vspace{-0.1in}

\subsubsection{Image-based \modelname Pre-training}
\label{sec:imagetraining}
Most VIS models~\cite{yang2019video,wu2021seqformer,cheng2021mask2former} are initialized from a model pretrained on the COCO instance segmentation dataset. To completely eliminate mask supervision, we pretrain our \modelname on COCO using only box supervision. We adopt the spatial consistency loss described in \cref{sec:spatial} on single frame to replace the original GT mask losses in Mask2Former~\cite{cheng2021mask2former}, while following the same image-based training setup on COCO. Thus, we provide two training settings in our experiments, one eliminates both image and video mask during training, while the other adopts weights pretrained with COCO mask annotations. In both cases, no video mask annotations are used.

\section{Experiments}

\subsection{Datasets}


\parsection{YTVIS 2019/2021}
We perform experiments on the large-scale YouTube-VIS~\cite{yang2019video} 2019 and 2021. YTVIS 2019 includes 2,883 videos of 131k annotated object instances belonging to 40 categories. 
To handle more complex cases, YTVIS 2021 updates  YTVIS 2019 with additional 794 videos for training and 129 videos for validation, including more tracklets with confusing motion trajectories.

\parsection{OVIS} 
We also train and evaluate on OVIS~\cite{qi2022occluded}, a VIS benchmark on occlusion learning. OVIS consists of instance masks covering 25 categories with 607, 140 and 154 videos for train, valid and test respectively.

\parsection{BDD100K MOTS}
We further report results of \modelname on the large-scale self-driving benchmark BDD100K~\cite{bdd100k} MOTS. The dataset annotates 154 videos (30,817 images) for training, 32 videos (6,475 images) for validation, and 37 videos (7,484 images) for testing.




\subsection{Implementation Details}
Our proposed approach only requires replacing the original video mask loss in state-of-the-art VIS methods. In particular, we adopt Mask2Former~\cite{cheng2021mask2former} and SeqFormer~\cite{wu2021seqformer} due to their excellent VIS results.
Unless specified, we kept all other training schedules and settings the same as in the original methods. 
For the \lossname, we set the patch size to 3$\times$3, search radius to 5 and $K=5$. We adopt the $L_2$ distance as the patch matching metric and set the matching threshold to 0.05.
On YTVIS 2019/2021, the Mask2Former based models are trained with AdamW~\cite{loshchilov2017decoupled} with learning rate  $10^{-4}$ and weight decay 0.05. The learning rate decays by 10 times at with a factor of 2/3. We set batch size to 16, and train 6k/8k iterations on YTVIS 2019/2021.
For experiments on OVIS and BDD100K, we adopted the COCO mask pretrained models by VITA and Unicorn.
For the sampled temporal tube at training, we use 5 frames with shuffling instead of 2 frames for better temporal regularization. 
The compared baselines are adjusted accordingly.
During testing, since there is no architecture modification, the inference of \modelname is the same to the baselines.
More details are in the Supp.\ file.

\subsection{Ablation Experiments}
\label{sec:ablation}
We perform detailed ablation studies for \modelname using ResNet-50 as backbone on the YTVIS 2019 \textit{val} set. We adopt the COCO box-pretrained model as initialization to eliminate all mask annotations from the training. Taking Mask2Former~\cite{cheng2021mask2former} as the base VIS method, we analyze the impact of individual proposed components. Moreover, we study several alternative solutions for temporal matching and influence of different hyper-parameters to TK-Loss.

\begin{table}[t]
	\caption{Different temporal matching schemes under the mask-free training setting on YTVIS2019 val. `Param' indicates whether the matching scheme brings extra model parameters.}
	\vspace{-0.1in}
	\centering
	\setlength\tabcolsep{4.0pt}
	\resizebox{1.0\linewidth}{!}{
		\begin{tabular}{l | c | c | c c c c}
			\toprule
			Temporal Matching Scheme & Param & AP & AP$_{50}$ & AP$_{75}$ & AR$_{1}$ & AR$_{10}$  \\
			\midrule
			Baseline (\textbf{No} Matching) & \xmark & 38.6 & 65.9 & 38.8 & 38.4 & 47.7 \\
			\midrule
			Flow-based Matching & \cmark & 40.2 & 66.3 & 41.9 & 40.5 & 49.1\\
			Temporal Deformable Matching & \cmark & 39.6 & 65.9 & 40.1 & 39.9 & 48.6 \\
			Learnable Pixel Matching & \cmark & 39.5 & 65.7 & 40.2 & 39.7 & 48.4\\
			Learnable Patch Matching & \cmark & 40.6 & 66.5 & 42.6 & 40.3 & 49.2 \\
			\midrule
			3D Pairwise Matching & \xmark & 39.4 & 65.0 & 41.7 & 40.2 & 48.0\\
			\midrule
			Temporal KNN-Matching (Ours) & \xmark & \textbf{42.5} & \textbf{66.8} & \textbf{45.7} & \textbf{41.2} & \textbf{51.2} \\
			
			\bottomrule
		\end{tabular}
	}
	\vspace{-0.25in}
	\label{tab:matching}
\end{table}

\parsection{Comparison on Temporal Matching Schemes}
Table~\ref{tab:matching} compares our Temporal KNN-Matching to four alternative frame-wise matching approaches for enforcing temporal mask consistency. \textbf{Flow-based Matching} employs the pretrained optical flow model RAFT~\cite{teed2020raft} to build pixel correspondence~\cite{liu2021weakly}. \textbf{Temporal Deformable Matching} adopts the temporal deformable kernels~\cite{tian2020tdan} to predict the pixel offsets between the target and alignment frame. Instead of using raw patches, \textbf{Learnable Pixel/Patch Matching} employs jointly learned deep pixel/patch embeddings via three learnable FC layers, which are then used to compute the affinities in a soft attention-like manner. \textbf{3D Pairwise Matching} directly extends $\mathcal{L}_\text{pair}$ designed for spatial images to the temporal dimension, where pairwise affinity loss is computed among pixels not only in within the frame but also across multiple frames. 

In Table~\ref{tab:matching}, compared to Flow-based Matching with one-to-one pixel correspondence, our paramter-free Temporal KNN-Matching with one-to-$K$ improves by about 2.3 AP. 
The prediction of flow-based models are not reliable in case of occlusions and homogeneous regions, and are also influenced by the gap between the training dataset and real-world video data.
For the above mentioned deformable and learnable matching schemes, since there are only weak bounding box labels during training, the temporal matching relation is learnt implicitly.
We empirically observe the instability during training with limited improvement under the mask-free training setting.
For direct generalization of $\mathcal{L}_\text{pair}$, it only leads to 0.8 mask AP performance improvement.
Despite the simplicity and efficiency of the TK-loss, it significantly improves VIS performance by 3.9 mask AP.





\begin{figure}[!t]
	\centering
	\vspace{-0.05in}
	\includegraphics[width=1.0\linewidth]{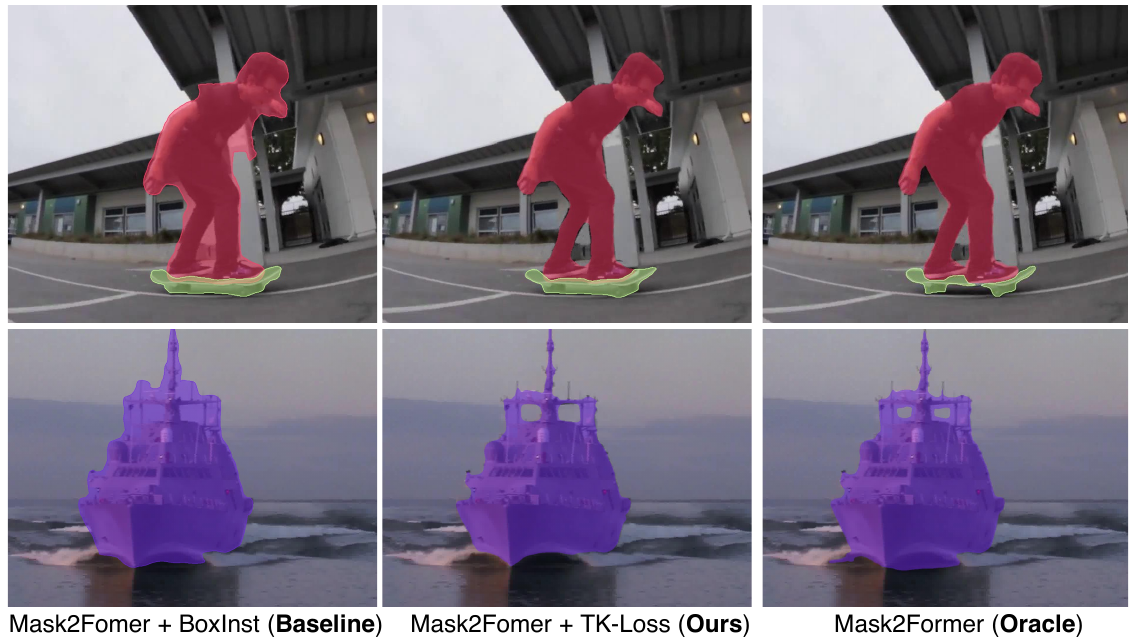}
	\vspace{-0.25in}
	\caption{Qualitative results comparison between using Spatial Pairwise loss~\cite{tian2021boxinst}, our TK-Loss, and Mask2Former (oracle) trained with GT video and image masks.}
	\label{fig:qualitative}
	\vspace{-0.1in}
\end{figure}

\parsection{Effect of \lossname}
\modelname is trained with joint spatio-temporal losses. In Table~\ref{tab:loss_component}, to evaluate the effectiveness of each loss component, we compare the performance of \modelname solely under the spatial pairwise loss~\cite{tian2021boxinst} or our proposed TK-Loss. Compared to the 2.0 mask AP improvement by the spatial pairwise loss, the TK-Loss substantially promotes the mask AP from 36.6 to 41.6, showing the advantage of our flexible one-to-$K$ patch correspondence design in leveraging temporal consistency. We show the VIS results in Figure~\ref{fig:qualitative} to visualize the effectiveness of each loss component.

\begin{table}[!t]
\footnotesize
    \vspace{-0.05in}
	\caption{Effect of the Spatial Pairwise loss and our \lossname on YTVIS2019 val.}
	\vspace{-0.1in}
	\centering
	\setlength\tabcolsep{5.0pt}
	\resizebox{1.0\linewidth}{!}{
		\begin{tabular}{c | c | c | l | c  c  c  c}
			\toprule
			Box Proj & Pairwise & TK-Loss & AP & AP$_{50}$ & AP$_{75}$ & AR$_{1}$ & AR$_{10}$  \\
			\midrule
			 \cmark & & & 36.6 & 66.5 & 36.2 & 37.1 & 45.0 \\
			 \cmark & \cmark & & 38.6$_{\uparrow2.0}$ & 65.9 & 38.8 & 38.4 & 47.7\\
			 \cmark & & \cmark & 41.6$_{\uparrow\textbf{5.0}}$ & \textbf{68.4} & 43.5 & 40.1 & 50.5\\
			 \cmark & \cmark & \cmark & \textbf{42.5}$_{\uparrow5.9}$ & 66.8 & \textbf{45.7} & \textbf{41.2} & \textbf{51.2} \\
			\bottomrule
		\end{tabular}
	}
	\vspace{-0.15in}
	\label{tab:loss_component}
\end{table}

\begin{table}[!t]
\footnotesize
	\begin{minipage}[t]{0.48\linewidth}
		\caption{Effect of using max $\textbf{K}$ matches on YTVIS2019 val.}
	\vspace{-0.12in}
	\centering
	\setlength\tabcolsep{5.0pt}
	\resizebox{0.98\linewidth}{!}{
		\begin{tabular}{l | c c c c}
			\toprule
			$\textbf{K}$ & AP & AP$_{50}$ & AP$_{75}$ & AR$_{1}$ \\
			\midrule
			1 & 40.8 & 65.8 & 44.1 & 40.3 \\
			3 & 41.9 & 66.9 & 45.1 & \textbf{41.9} \\
			\textbf{5} & \textbf{42.5} & 66.8 & \textbf{45.7} & 41.2 \\
			7 & 42.3 & \textbf{67.1} & 44.6 & 40.6 \\
			\bottomrule
		\end{tabular}
	}
		\label{tab:k_influence}
	\end{minipage}
	\hfill
	\begin{minipage}[t]{0.48\linewidth}
		\caption{Patch Matching metrics comparison. NCC is Norm.\ Cross-Correlation.}
	\vspace{-0.12in}
	\centering
	\setlength\tabcolsep{4.0pt}
	\resizebox{0.99\linewidth}{!}{
		\begin{tabular}{c | c c c c}
			\toprule
			Metric & AP & AP$_{50}$ & AP$_{75}$ & AR$_{1}$ \\
			\midrule
			NCC & 41.7 & 66.7 & 43.4 & \textbf{41.4} \\
			L$_1$ & 41.2 & 66.2 & 43.6 & 40.3  \\
			L$_2$ & \textbf{42.5} & \textbf{66.8} & \textbf{45.7} & 41.2 \\
			\bottomrule
		\end{tabular}
	}
		\label{tab:patch_metric}
	\end{minipage}
	\vspace{-0.1in}
\end{table}

\begin{table}[!t]
\footnotesize
	\begin{minipage}[t]{0.48\linewidth}
		\caption{Impact of search radius $\textbf{R}$ on YTVIS2019 val.}
	\vspace{-0.12in}
	\centering
	\setlength\tabcolsep{5.0pt}
	\resizebox{0.98\linewidth}{!}{
		\begin{tabular}{l | c c c c }
			\toprule
			$R$ & AP & AP$_{50}$ & AP$_{75}$ & AR$_{1}$  \\
			\midrule
			1 & 39.6 & 65.7 & 40.2 & 39.7 \\
			3 & 41.3 & 66.6 & 43.0 & 40.3 \\
			\textbf{5} & \textbf{42.5} & 66.8 & \textbf{45.7} & \textbf{41.2} \\
			7 & 42.3 & \textbf{67.1} & 44.6 & 40.6 \\
			\bottomrule
		\end{tabular}
	}
		\label{tab:radius}
	\end{minipage}
	\hfill
	\begin{minipage}[t]{0.48\linewidth}
		\caption{Influence of patch size $\textbf{N}$ on YTVIS2019 val.}
	\vspace{-0.12in}
	\centering
	\setlength\tabcolsep{5.0pt}
	\resizebox{0.98\linewidth}{!}{
		\begin{tabular}{l | c c c c }
			\toprule
			$N$ & AP & AP$_{50}$ & AP$_{75}$ & AR$_{1}$  \\
			\midrule
			1 & 40.1 & 65.2 & 42.2 & 40.0 \\
			\textbf{3} & \textbf{42.5} & 66.8 & \textbf{45.7} & 41.2 \\
			5 & 42.1 & \textbf{68.7} & 44.4 & \textbf{42.1} \\
			7 & 41.5 & 66.3 & 44.8 & 41.3\\
			\bottomrule
		\end{tabular}
	}
		\label{tab:patch_size}
	\end{minipage}
	\vspace{-0.2in}
\end{table}

\parsection{Analysis of \lossname}
In Table~\ref{tab:k_influence}, we study the influence of $\textbf{K}$, the maximum number of matches selected in \lossname. The best result is obtained for $K=5$, while $K=1$ only allows for the one-to-one and no-match cases. The improvement from 40.8 mask AP to 42.5 mask AP reveals the benefit brought by the flexible one-to-many correspondence design. 
We also analyze the matching metric in Table~\ref{tab:patch_metric}, search radius in Table~\ref{tab:radius}, and patch size in Table~\ref{tab:patch_size} (see Sec.~\ref{sec:tkloss} for details). When patch size $N$ is increased from 1 to 3 in Table~\ref{tab:patch_size}, the performance of \modelname is improved by 2.4 AP, validating the importance of patch structure in robust matching.

\parsection{Effect of the Cyclic Tube Connection} 
We compare three frame-wise tube connection schemes (Figure~\ref{fig:tube_connect}) for the TK-Loss in Table~\ref{tab:tube_conn}. While dense connection brings forth the best performance, it doubles the training memory with minor improvement compared to Cyclic connection.
Comparing to Sequential connection, our Cyclic connection benefits from long-range consistency, improving the performance of 0.6 mask AP with an affordable memory cost growth. 


\begin{table}[!h]
\footnotesize
    \vspace{-0.15in}
	\caption{Comparison of the tube connection schemes, illustrated in \cref{fig:tube_connect}. The tube consist of 5 frames. `Mem' denotes the memory consumption per sampled video by the TK-Loss during training.}
	\vspace{-0.1in}
	\centering
	\setlength\tabcolsep{3.0pt}
	\resizebox{1.0\linewidth}{!}{
		\begin{tabular}{l | c | c | c c c c}
			\toprule
			Tube Connect & Connect Num. & Mem (MB) & AP & AP$_{50}$ & AP$_{75}$ & AR$_{1}$  \\
			\midrule
			Dense & 10 & 1526 & 42.7 & 68.0 & 44.3 & 41.5 \\
			Sequential & 4 & 631 & 41.9 & 66.5 & 44.7 & 41.2 \\
			Cyclic (Ours) & 5 & 773 & 42.5 & 66.8 & 45.7 & 41.2 \\
			\bottomrule
		\end{tabular}
	}
	\vspace{-0.1in}
	\label{tab:tube_conn}
\end{table}




\parsection{Comparison on Sequence Matching Functions} Besides TK-Loss, we analyze the influence of sequence matching cost functions for transformer-based VIS methods under the mask-free setting.
In Table~\ref{tab:set_match}, we identify the substantial advantage of Spatio-temporal box-mask matching over frame-wise cost averaging~\cite{wu2021seqformer,wang2021end}.
As discussed in Sec.~\ref{sec:integrate}, we achieve further gain by removing the object size imbalanced cross-entropy cost computation. 

\begin{figure}[!t]
	\centering
	\includegraphics[width=1.0\linewidth]{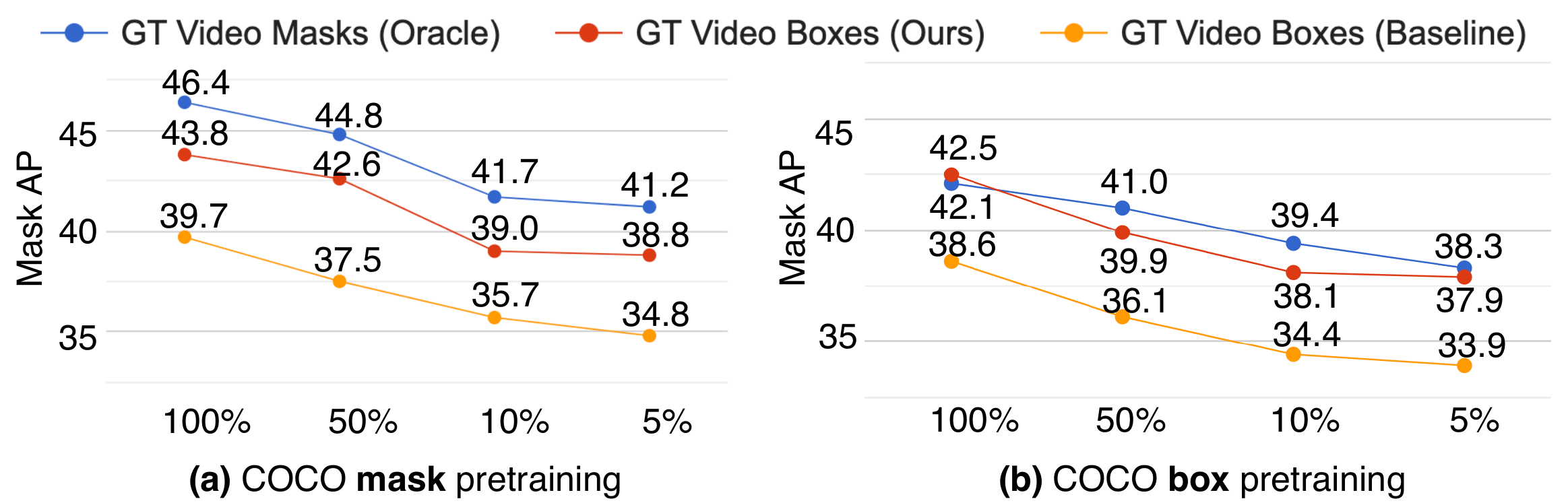}
	\vspace{-0.25in}
	\caption{Results on YTVIS 2019 val with various percentages of the YTVIS training data. Baseline denotes Mask2Former~\cite{cheng2021mask2former} trained with GT video boxes using BoxInst~\cite{tian2021boxinst}, while Oracle denotes Mask2Former trained with GT video masks.}
	\label{fig:percent}
	\vspace{-0.12in}
\end{figure}

\begin{table}[!t]
\footnotesize
    \vspace{-0.02in}
	\caption{Comparison of Set Matching Cost Functions on YTVIS2019 val. ST-BoxMask denotes our Spatio-temporal Box Mask matching. \textbf{w/o} CE denotes removing cross-entropy cost.}
	\vspace{-0.12in}
	\centering
	\setlength{\tabcolsep}{6.0pt}
	\resizebox{0.95\linewidth}{!}{
		\begin{tabular}{c | c c c c c }
			\toprule
			Matching Cost Function & AP & AP$_{50}$ & AP$_{75}$ & AR$_{1}$ & AR$_{10}$  \\
			\midrule
			 Frame-wise Averaging~\cite{wu2021seqformer,wang2021end} & 37.6 & 64.2 & 39.5 & 37.5 & 45.7 \\
			 ST-BoxMask & 40.8 & \textbf{67.8} & 42.2 & 40.0 & 49.2 \\
			 ST-BoxMask \textbf{w/o} CE & \textbf{42.5} & 66.8 & \textbf{45.7} & \textbf{41.2} & \textbf{51.2} \\
			\bottomrule
		\end{tabular}
	}
	\vspace{-0.25in}
	\label{tab:set_match}
\end{table}

\parsection{Training on Various Amounts of Data} To further study label-efficient VIS, we validate the effect of \modelname under various percentages of the YTVIS 2019 training data. We uniformly sample frames and their labels for each video, and set the minimum sampled number of frames to 1. Figure~\ref{fig:percent} presents the experimental results, which shows the consistent large improvement (over 3.0 AP) brought by our TK-Loss under various amount of data. In particular, we note that our approach with 50\% data even outperforms the fully-supervised model with 10\% training data. 


\begin{table}[t!]
    \vspace{-0.15in}
    \newcommand{\hsp}{~~}%
	\centering%
		\caption{Comparison on YTVIS 2019. I: using COCO mask pretrained model as initialization. V: using YTVIS video masks during training. $\ast$: using pseudo mask from COCO images for joint training~\cite{wu2021seqformer}. M2F: Mask2Former~\cite{cheng2021mask2former}, SeqF: SeqFormer~\cite{wu2021seqformer}.}
	\setlength{\tabcolsep}{2.4pt}
	\vspace{-0.13in}
	\resizebox{0.48\textwidth}{!}{%
	\begin{tabular}{lclcccccc}
					\toprule
					Method & Mask & Back- & AP & AP$_{50}$ & AP$_{75}$ & AR$_{1}$ & AR$_{10}$ \\
					& ann. & bone & & & & & \\
					\midrule
					\underline{\textit{Fully-supervised}:} &  & & &  & &  \\
				    PCAN~\cite{pcan} & I+V & R50 & 36.1 & 54.9 & 39.4 & 36.3 & 41.6 \\
				    EfficientVIS~\cite{wu2022efficient} & I+V & R50 & 37.9 & 59.7 & 43.0 & 40.3 & 46.6 \\
					InsPro~\cite{inspro} & I+V & R50 & 40.2 & 62.9 & 43.1 & 37.6 & 44.5 \\
					IFC~\cite{hwang2021video} & I+V & R50 & 42.8 & 65.8 & 46.8 & 43.8 & 51.2 \\
					VMT$^{\ast}$~\cite{vmt} & I+V & R50  & 47.9 & - & 52.0 & 45.8 & - \\
					SeqF$^{\ast}$~\cite{wu2021seqformer} & I+V & R50  & 47.4 & 69.8 & 51.8 & 45.5 & 54.8 \\
					M2F & I+V & R50  & 46.4 & 68.0 & 50.0 & - & -\\
					M2F$^{\ast}$ & I+V & R50  & 47.8 & 69.2 & 52.7 & 46.2 & 56.6 \\
					\midrule
					\underline{\textit{Prev.\ Weakly-supervised}:} &  & & &  & &  \\
					FlowIRN~\cite{liu2021weakly} & - & R50 & 10.5 & 27.2 & 6.2 & 12.3 & 13.6 \\ 
					SOLO-Track~\cite{fu2021learning} & I & R50 & 30.6 & 50.7 & 33.5 & 31.6 & 37.1 \\ 
					\midrule
					\underline{\textit{Mask-free}:} &  & & &  & &  \\
					M2F + Flow Consist~\cite{teed2020raft} & - & R50 & 40.2 & 66.3 & 41.9 & 40.5 & 49.1 \\
					M2F + BoxInst~\cite{tian2021boxinst} & - & R50 & 38.6 & 64.2 & 38.5 & 38.0 & 46.8 \\
					M2F + \textbf{\modelname} & - & R50 & 42.5$_{\uparrow\textbf{3.9}}$ & 66.8 & 45.7 & 41.2 & 51.2 \\
					M2F + \textbf{\modelname} & I & R50  & 43.8$_{\uparrow\textbf{5.2}}$ & 70.7 & 46.9 & 41.5 & 52.3 \\
					M2F + \textbf{\modelname}$^{\ast}$  & I & R50  & \textbf{46.6}$_{\uparrow\textbf{8.0}}$ & \textbf{72.5} & \textbf{49.7} & \textbf{44.9} & \textbf{55.7} \\
					\specialrule{.1em}{.05em}{.05em} 
					\underline{\textit{Fully-supervised}:} &  & & &  & &  \\
					M2F & V & R101 & 45.6 & 72.6 & 48.9 & 44.3 & 54.5 \\
					M2F & I+V & R101 & 49.2 & 72.8 & 54.2 & - & - \\
					M2F$^{\ast}$ & I+V & R101 & 49.8 & 73.6 & 55.4 & 48.0 & 58.0 \\
					SeqF$^{\ast}$ & I+V & R101  & 49.0 & 71.1 & 55.7 & 46.8 & 56.9 \\
					\midrule
					\underline{\textit{Mask-free}:} &  & & &  & &  \\
					M2F + BoxInst~\cite{tian2021boxinst} & - & R101 & 40.8 & 67.8 & 42.2 & 40.0 & 49.2 \\
					M2F + \textbf{\modelname} & - & R101 & 45.8$_{\uparrow\textbf{5.0}}$ & 70.8 & 48.6 & \textbf{45.3} & 55.2 \\
					M2F + \textbf{\modelname} & I & R101  & 47.3$_{\uparrow\textbf{6.5}}$ & \textbf{75.4} & 49.9 & 44.6 & 55.2 \\
					M2F + \textbf{\modelname}$^{\ast}$  & I & R101 & \textbf{48.9}$_{\uparrow\textbf{8.1}}$ & 74.9 & \textbf{54.7} & 44.9 & 57.0 \\
					SeqF + \textbf{\modelname}$^{\ast}$ & I & R101 & 48.6 & 74.0 & 52.2 & 45.9 & \textbf{57.2} \\
					\specialrule{.1em}{.05em}{.05em} 
					\underline{\textit{Fully-supervised}:} &  & & &  & &  \\
					M2F & I+V & SwinL & 60.4 & 84.4 & 67.0 & - & - \\
					\midrule
					\underline{\textit{Mask-free}:} &  & & &  & &  \\
					M2F + BoxInst~\cite{tian2021boxinst} & - & SwinL & 49.8 & 73.2 & 55.5 & 48.2 & 58.1 \\
					M2F + \textbf{\modelname} & - & SwinL & 54.3$_{\uparrow\textbf{4.5}}$ & 82.6 & 61.1 & 50.2 & 61.3 \\
					M2F + \textbf{\modelname}$^{\ast}$ & I & SwinL & \textbf{55.3}$_{\uparrow\textbf{5.5}}$ & 82.5 & 60.8 & 50.7 & 62.2 \\
					\bottomrule
			\end{tabular}}
		\label{table:vis19}
		\vspace{-0.1in}
	\end{table}

\subsection{Comparison with State-of-the-art Methods}

We compare \modelname with the state-of-the-art fully/weakly supervised methods on benchmarks YTVIS 2019/2021, OVIS and BDD100K MOTS. We integrate \modelname on four representative methods~\cite{cheng2021mask2former,wu2021seqformer,heo2022vita,unicorn}, attaining consistent large gains over the strong baselines.

\parsection{YTVIS 2019/2021} 
Table~\ref{table:vis19} compares the performance on YTVIS 2019. Using R50/R101 as backbone and with the same training setting, \modelname achieves 42.5/45.8 AP, improving 3.9/5.0 AP over the strong baseline adopting BoxInst~\cite{tian2021boxinst} losses. \modelname, \textbf{without} any mask labels, even \textit{significantly outperforms some recent fully-supervised methods} such as EfficientVIS~\cite{wu2022efficient} and InsPro~\cite{inspro}. On R50/R101/Swin-L, our~\modelname consistently attains over 91\% of its fully-supervised counterpart trained with both GT image and video masks. We also observe similar larger performance growth over the baseline on YTVIS 2021 in Table~\ref{table:vis21}.
The excellent performance substantially narrows the performance gap between fully-supervised and weakly-supervised VIS.







\begin{table}[t!]
\vspace{-0.15in}
\centering%
\footnotesize
\caption{Comparison on YTVIS 2021. Refer to Table~\ref{table:vis19} for the symbol abbreviations.} 
\vspace{-0.08in}
\setlength{\tabcolsep}{2.4pt}
\vspace{-0.05in}
\resizebox{0.48\textwidth}{!}{%
\begin{tabular}{lclcccccc}
					\toprule
					Method & Mask & Back- & AP & AP$_{50}$ & AP$_{75}$ & AR$_{1}$ & AR$_{10}$ \\
					& ann. & bone & & & & & \\
					\midrule
					\underline{\textit{Fully-supervised}:} &  & & &  & &  \\
					MaskTrack~\cite{yang2019video} & I+V & R50 & 28.6 & 48.9 & 29.6 & 26.5 & 33.8 \\
					IFC~\cite{hwang2021video} & I+V & R50 & 36.6 & 57.9 & 39.3 & - & - \\
					SeqF$^{\ast}$~\cite{wu2021seqformer} & I+V & R50  & 40.5 & 62.4 & 43.7 & 36.1 & 48.1 \\
					M2F & I+V & R50  & 40.6 & 60.9 & 41.8 & - & -\\
					\midrule
					\underline{\textit{Mask-free}:} &  & & &  & &  \\
					M2F + BoxInst~\cite{tian2021boxinst} & - & R50 & 32.1 & 52.8 & 34.4 & 31.0 & 38.1 \\
					M2F + \textbf{\modelname} & - & R50 & 36.2$_{\uparrow\textbf{4.1}}$ & 60.8 & 39.2 & 34.6 & 45.6 \\
					M2F + \textbf{\modelname} & I & R50  & 37.2$_{\uparrow\textbf{5.1}}$ & 61.9 & 40.3 & 35.3 & 46.1 \\
					M2F + \textbf{\modelname}$^{\ast}$ & I & R50  & \textbf{40.9}$_{\uparrow\textbf{8.8}}$ & \textbf{65.8} & \textbf{43.3} & \textbf{37.1} & \textbf{50.5} \\
					\specialrule{.1em}{.05em}{.05em}  
					\underline{\textit{Fully-supervised}:} &  & & &  & &  \\
					M2F & I+V & R101  & 42.4 & 65.9 & 45.8 & - & -\\
					\midrule
					\underline{\textit{Mask-free}:} &  & & &  & &  \\
					M2F + BoxInst~\cite{tian2021boxinst} & - & R101 & 33.3 & 55.2 & 32.5 & 32.1 & 41.9 \\
					M2F + \textbf{\modelname} & - & R101 & 37.3$_{\uparrow\textbf{4.0}}$ & 61.6 & 39.4 & 34.1 & 45.6 \\
					M2F + \textbf{\modelname} & I & R101  & 38.2$_{\uparrow\textbf{4.9}}$ & 62.4 & 40.0 & 34.9 & 46.2 \\
					M2F + \textbf{\modelname}$^{\ast}$ & I & R101  & \textbf{41.6}$_{\uparrow\textbf{8.3}}$ & \textbf{66.2} & \textbf{44.8} & \textbf{36.3} & \textbf{49.2} \\
					\bottomrule
\end{tabular}}
\label{table:vis21}
\vspace{-3mm}
\end{table}
	
	\begin{table}[t!]
        \footnotesize
	\centering%
		\caption{State-of-the-art comparison on the OVIS using R50.}
		\setlength{\tabcolsep}{3.8pt}
		\vspace{-0.1in}
	\begin{tabular}{llccccc}
					\toprule
					Method & AP & AP$_{50}$ & AP$_{75}$ & AR$_{1}$ & AR$_{10}$ \\
					\midrule
					\underline{\textit{Fully-supervised}:} &  & & &  & \\
					CrossVIS~\cite{Yang_2021_ICCV} & 14.9 & 32.7 & 12.1 & 10.3 & 19.8 \\
					Mask2Former~\cite{cheng2021mask2former} & 17.3 & 37.3 & 15.1 & 10.5 & 23.5 \\ 
					VMT~\cite{vmt} & 16.9 & 36.4  & 13.7 & 10.4 & 22.7 \\ 
					VITA~\cite{heo2022vita}& 19.6 & 41.2 & 17.4 & 11.7 & 26.0 \\
					\midrule
					\underline{\textit{Video Mask-free}:} &  & & &  & \\
					VITA~\cite{heo2022vita} + BoxInst~\cite{tian2021boxinst} & 12.1 & 28.3 & 10.2 & 8.8 & 17.9 \\
					VITA~\cite{heo2022vita} + \textbf{\modelname} & \textbf{15.7}$_{\uparrow\textbf{3.6}}$ & \textbf{35.1} & \textbf{13.1} & \textbf{10.1} & \textbf{20.4} \\
					\bottomrule
			\end{tabular}
		\label{table:ovis}
		\vspace{-2.5mm}
	\end{table}
 
\begin{table}[t!]
        \scriptsize
	\centering%
		\caption{State-of-the-art comparison on the BDD100K segmentation tracking validation set.}
		\vspace{-0.1in}
		\setlength{\tabcolsep}{2pt}
				\begin{tabular}{lccccc}
					\toprule
					Method & mMOTSA$\uparrow$ & mMOTSP$\uparrow$ & mIDF$\uparrow$ & ID sw.$\downarrow$  & mAP$\uparrow$ \\
					\midrule
					\underline{\textit{Fully-supervised}:} &  & & &  & \\
					STEm-Seg~\cite{Athar_Mahadevan20ECCV} & 12.2 & 58.2 & 25.4 & 8732 & 21.8 \\
					QDTrack-mots-fix~\cite{qdtrack} & 23.5 & 66.3 & 44.5 & 973 & 25.5 \\
					PCAN~\cite{pcan} & 27.4 & 66.7 & 45.1 & 876 & 26.6 \\
					Unicorn~\cite{unicorn} & 29.6 & 67.7 & 44.2 & 1731 & 32.1 \\
					\midrule
					\underline{\textit{Video Mask-free}:} &  & & &  & \\
					Unicorn~\cite{unicorn} + BoxInst~\cite{tian2021boxinst} & 18.9 & 58.7 & 36.3 & 3298 & 22.1 \\
					Unicorn~\cite{unicorn} + \textbf{\modelname} & \textbf{23.8}$_{\uparrow\textbf{4.9}}$ & \textbf{66.7} & \textbf{44.9} & \textbf{2086} & \textbf{24.8} \\
				   \bottomrule
			\end{tabular}
		\label{table:bdd}
		\vspace{-5mm}
\end{table}

\parsection{OVIS} We also conduct experiments on OVIS in Table~\ref{table:ovis} using R50 as backbone. We integrate \modelname with VITA~\cite{heo2022vita}, promoting the baseline performance from 12.1 to 15.7 under the mask-free training setting. 

\parsection{BDD100K MOTS} Table~\ref{table:bdd} further validates our approach on BDD100K MOTS. Integrated with Unicorn~\cite{unicorn}, \modelname achieves 23.8 mMOTSA by improving over 4.9 points over the strong baseline and thus surpassing the fully-supervised QDTrack-mots~\cite{qdtrack}. The consistent large gains on four benchmarks and four base VIS methods validates the generalizability of our \modelname. 


\section{Conclusion} 
\modelname is the first competitive VIS method that does not need \emph{any} mask annotations during training. The strong results lead to a remarkable conclusion: mask labels are not a necessity for high-performing VIS. 
Our key component is the unsupervised \lossname, which replaces the conventional video masks losses by leveraging temporal mask consistency constraints. Our approach greatly reduces the long-standing gap between fully-supervised and weakly-supervised VIS on four large-scale benchmarks.
\modelname thus opens up many opportunities for researchers and practitioners for label-efficient VIS. 

\section{Appendix}
In this supplementary material, we first conduct additional experiment analysis of our~\lossname (TK-Loss) in~Section~\ref{sec:supp_exp}. Then, we present visualization of temporal matching correspondence and compute its approximate accuracy in Section~\ref{sec:supp_vis}.
We further show more qualitative VIS results analysis (including failure cases) in Section~\ref{sec:supp_quali}.
Finally, we provide \modelname algorithm pseudocode and more implementation details in~Section~\ref{sec:supp_details}.
Please refer to our project page for extensive MaskFreeVIS video results.


\subsection{Supplementary Experiments}
\label{sec:supp_exp}

\parsection{Patch \vs Pixel in TK-Loss}
Extending Table 4 in the paper, in Table~\ref{tab:patch_pixel}, we further compare the results of image patch \vs single pixels under different max $K$ values during temporal matching.
The one-to-$K$ correspondence produces gains in both pixel and patch matching manners, while the improvement on patch matching is much more obvious.

\begin{table}[!h]
    \vspace{-0.1in}
	\caption{Patch \vs Pixel in one-to-$\textbf{K}$ patch correspondence on YouTube-VIS 2019.}
	\vspace{-0.1in}
	\centering
	\setlength\tabcolsep{8.5pt}
	\resizebox{1.0\linewidth}{!}{
		\begin{tabular}{l | c | c | c c c c c}
			\toprule
			$\textbf{K}$ & Pixel & Patch & AP & AP$_{50}$ & AP$_{75}$ & AR$_{1}$ & AR$_{10}$  \\
			\midrule
			1 & \cmark & & 39.1 & 64.8 & 41.7 & 39.8 & 47.8 \\
			1 & & \cmark & 40.8 & 65.8 & 44.1 & 40.3 & 48.9 \\
			\midrule
			3 & \cmark & & 39.8 & 65.9 & 40.6 & 39.6 & 48.2 \\
			3 & & \cmark & 41.9 & 66.9 & 45.1 & \textbf{41.9} & 50.3 \\
			\midrule
			5 & \cmark & & 40.1 & 65.2 & 42.2 & 40.0 & 48.2 \\
			\textbf{5} & & \cmark & \textbf{42.5} & 66.8 & \textbf{45.7} & 41.2 & \textbf{51.2}\\
			\midrule
			7 & \cmark & & 39.6 & 64.9 & 41.0 & 39.8 & 48.5 \\
			7 & & \cmark & 42.3 & \textbf{67.1} & 44.6 & 40.6 & 50.7 \\
			\bottomrule
		\end{tabular}
	}
	\vspace{-0.1in}
	\label{tab:patch_pixel}
\end{table}

\parsection{Influence of Tube Length}
During model training, we sample a temporal tube from the video. 
We study the influence of the sampled tube lengths in Table~\ref{table:tube_len_supp}, and observe that the performing of \modelname saturates at temporal tube length 5. For even longer temporal tube, different from~\cite{pcan}, the temporal correlation between the beginning frame and ending frame (two temporally most distant frame) is weak to find sufficient patch correspondence.

\begin{table}[!h]
    \centering%
    \setlength\tabcolsep{10.0pt}
	\caption{Results of varying \textbf{Tube Length} during training for TK-Loss on YouTube-VIS 2019. Tube length 1 denotes model training with \textbf{only} spatial losses in BoxInst~\cite{tian2021boxinst}.}
	\vspace{-0.1in}
			\resizebox{1.0\linewidth}{!}{%
				\begin{tabular}{c | c c c c c }
					\toprule
					Tube Length & AP & AP$_{50}$ & AP$_{75}$ & AR$_{1}$ & AR$_{10}$ \\
					\midrule
					1 & 38.3 & 65.4 & 38.5 & 38.0 & 47.4 \\
					3 & 42.1 & 66.4 & 44.9 & 41.0 & 50.8 \\
					5 & \textbf{42.5} & 66.8 & \textbf{45.7} & 41.2 & \textbf{51.2} \\
					7 & 42.5 & \textbf{67.5} & 45.2 & \textbf{41.3} & 51.1 \\
					\bottomrule
				\end{tabular}
			}
	\label{table:tube_len_supp}%
	\vspace{-1mm}
\end{table}

\begin{figure}[!t]
	\centering
	\includegraphics[width=1.0\linewidth]{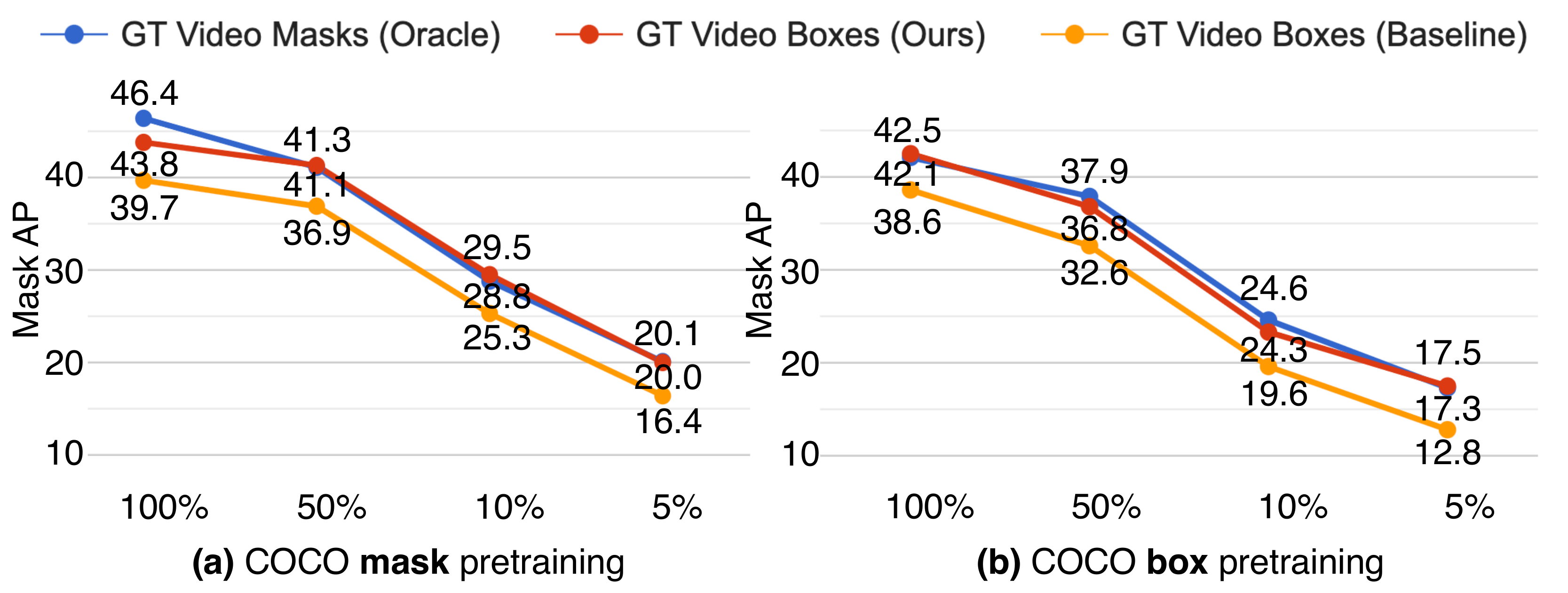}
	\vspace{-0.25in}
	\caption{Results on YTVIS 2019 val with various percentages of the YTVIS training data, by \textit{directly sampling different numbers of videos} from the YTVIS training set. Baseline denotes Mask2Former~\cite{cheng2021mask2former} trained with GT video boxes using BoxInst~\cite{tian2021boxinst}, while Oracle denotes the fully supervised Mask2Former trained with GT video masks.}
	\label{fig:percent1}
	\vspace{-0.01in}
\end{figure}

\parsection{Additional Results on Various Amount of YTVIS Data}
For experiments in Figure 6 of the paper, we sample different portions (in percents) of YTVIS data by uniformly sampling frames per video.
In Figure~\ref{fig:percent1}, we  experiment with another video sampling strategy by directly sampling different numbers of videos from the YTVIS training set. 
Our \modelname consistently attains large improvements (over 3.5 mask AP) over the baseline in both the COCO mask and box pretraining settings, with performance on par with the oracle Mask2Former in data-efficient settings.

\begin{figure*}[!t]
	\centering
	\includegraphics[width=1.0\linewidth]{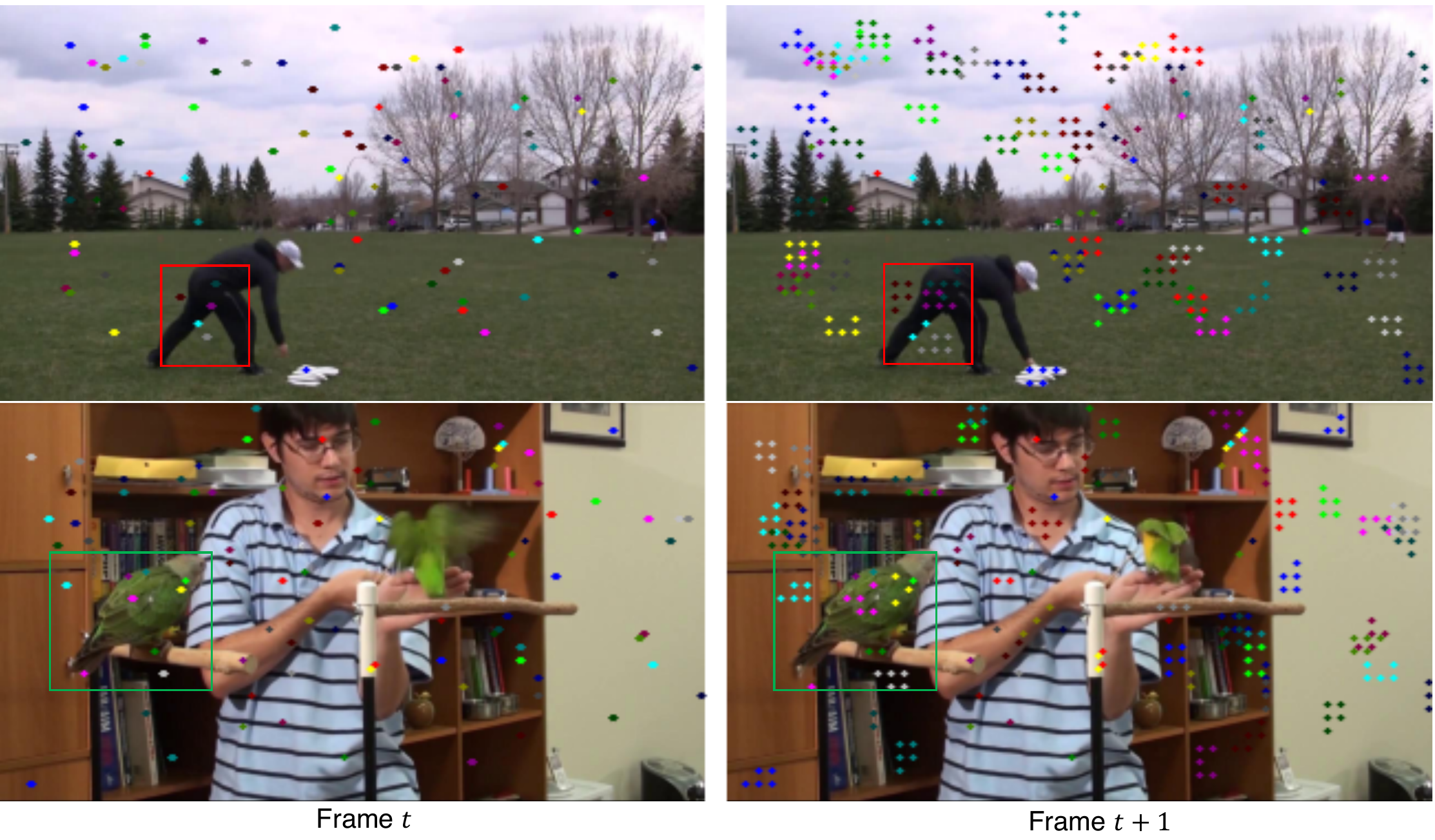}
	\vspace{-0.25in}
	\caption{Visualization of the temporal correspondence in TK-Loss. We randomly sample 100 patch center points from Frame $t$, and draw its temporally matched patch center points in Frame $t$+1. Matches are shown in the same color, and should have consistent instance mask label. Taking patch center points near the left leg of the man (inside the red box, 1st row in Frame $t$) as an example, the matches in Frame $t$+1 consistently belong to the same foreground (leg) / background (grass) region. Best viewed in color. }
	\label{fig:corres_t}
	\vspace{-0.01in}
\end{figure*}

\parsection{Image-based Pretraining Results on COCO}
In Table~\ref{table:coco_res}, we report the performance on COCO of image-pretrained Mask2Former networks used as initial weights for our approach. The mask-free version employs the spatial losses of BoxInst~\cite{tian2021boxinst}. We also show the corresponding VIS results on YTVIS 2019 by taking these image-pretrained models as initialization for our approach.
Compared to the fully-supervised Mask2Former on COCO, the box-training process eliminates the image masks usage and obtaining a lower performance (over 10.0 AP) in image mask AP on COCO. 
However, even initialized from this low-performing image-pretrained models, our \modelname using the proposed TK-Loss still greatly reduces the gap between fully-supervised and weakly-supervised VIS models as shown in the rightmost column of the Table~\ref{table:coco_res}.

\begin{table}[!h]
    \centering%
    \setlength\tabcolsep{6.5pt}
	\caption{Results of image-based pretrained Mask2Former (M2F)~\cite{cheng2022imgmask2former} on COCO \textit{val} and the corresponding video results on YTVIS 2019 by taking the image-pretrained one as initialization. M2F + BoxInst is mask-free, which is used to initialize \modelname, while image-based M2F (Oracle) is to initialize video-based M2F (Oracle). Oracle denotes training with GT image or video masks.}
	\vspace{-0.1in}
			\resizebox{1.0\linewidth}{!}{%
				\begin{tabular}{c | c  c| c c }
					\toprule
					Backbone & Image Method  & Image AP & VIS Method & Video AP \\
					\midrule
					R50 & M2F + BoxInst & 32.6 & \modelname & 42.5  \\
					R50 & M2F (Oracle) & 43.7 & M2F (Oracle) & 46.4  \\
					\midrule
					R101 & M2F + BoxInst  & 34.5 & \modelname & 45.8 \\
					R101 & M2F (Oracle) & 44.2 & M2F (Oracle) & 49.2 \\
					\midrule
					SwinL & M2F + BoxInst  & 40.3 & \modelname & 54.3 \\
					SwinL & M2F (Oracle) & 50.1 & M2F (Oracle) & 60.4  \\
					\bottomrule
				\end{tabular}
			}
	\label{table:coco_res}%
	\vspace{-3mm}
\end{table}

\begin{figure*}[!t]
	\centering
	\vspace{-0.01in}
	\includegraphics[width=1.0\linewidth]{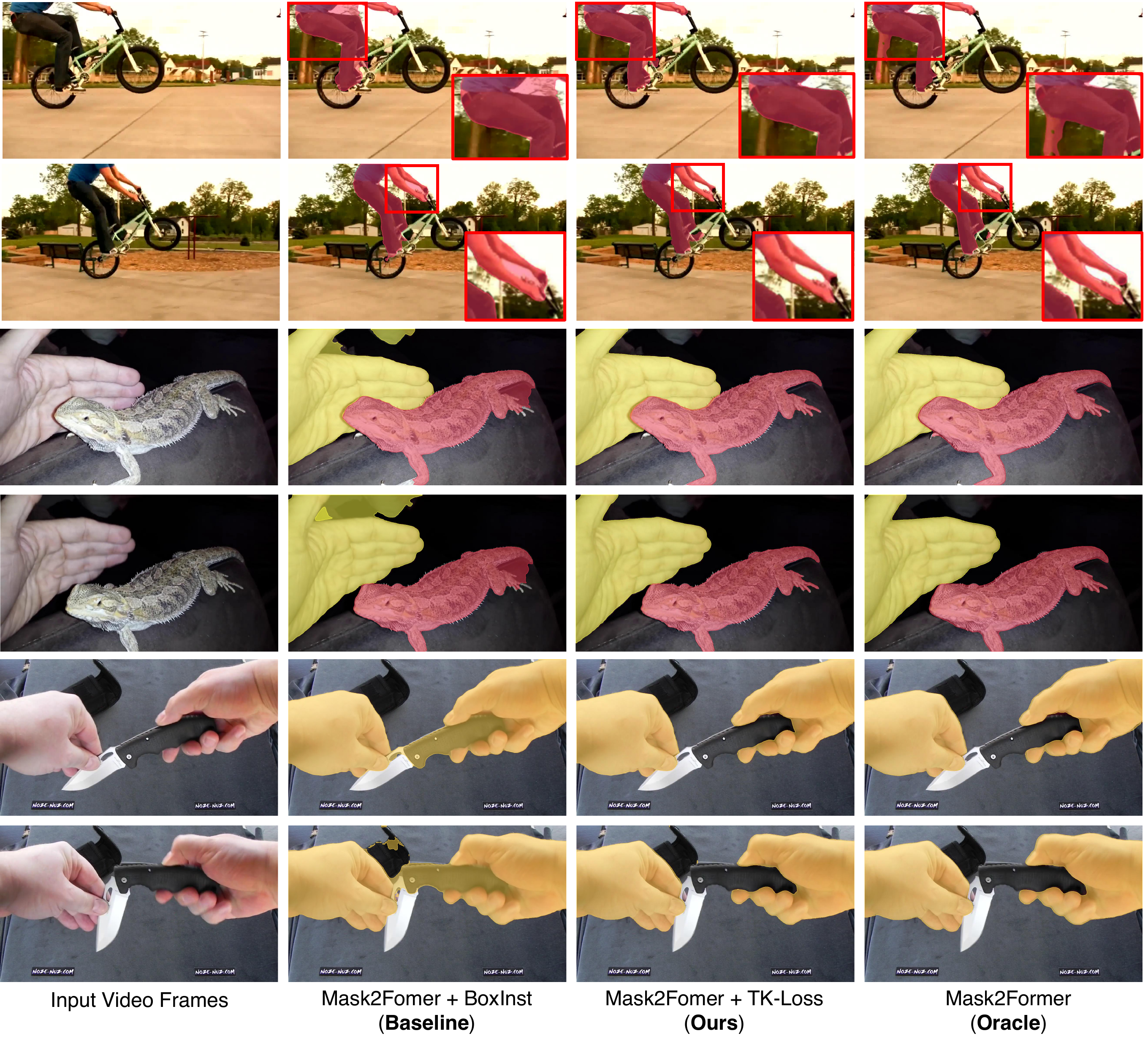}
	\vspace{-0.01in}
	\caption{Qualitative video instance segmentation results comparison between Mask2Former using Spatial Pairwise loss of BoxInst~\cite{tian2021boxinst} (Baseline), our proposed TK-Loss (Ours), and Mask2Former (Oracle) trained with GT video and image masks.}
	\label{fig:qualitative_supp}
	\vspace{-0.1in}
\end{figure*}

\begin{figure*}[!h]
	\centering
	\vspace{-0.01in}
	\includegraphics[width=1.0\linewidth]{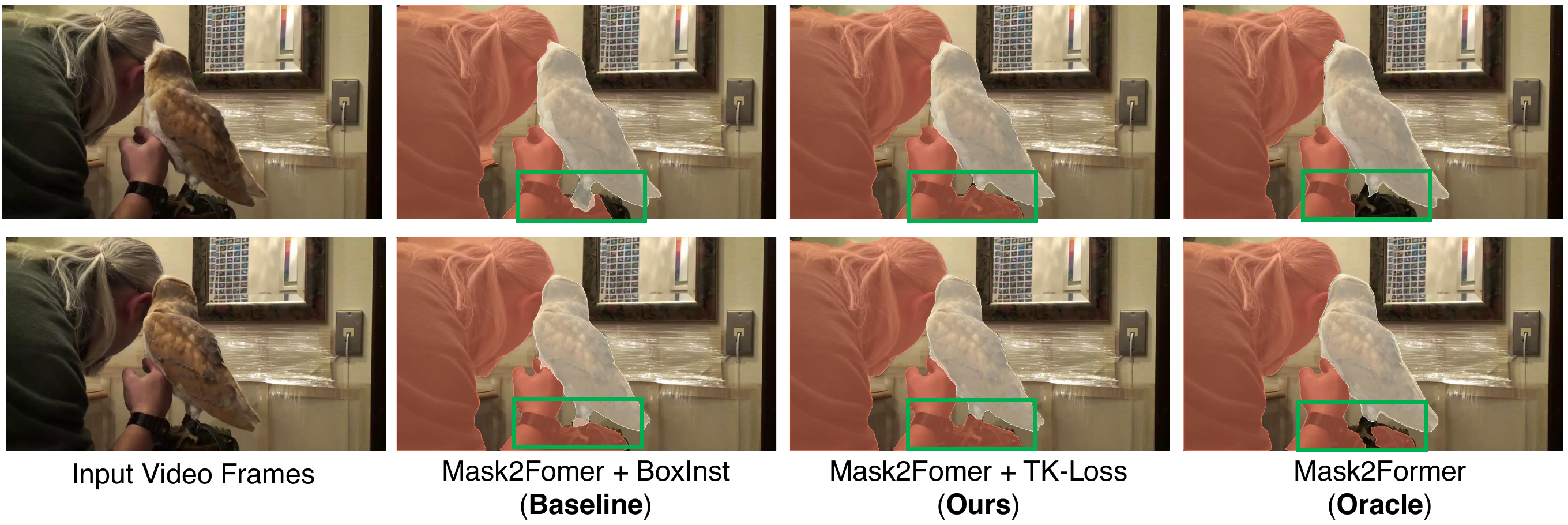}
	\vspace{-0.01in}
	\caption{One typical failure case of our \modelname. The neighboring hand watch and shelf belong to the same black color, and continuously closing to each other with no sufficient motion information for delineating these two objects.}
	\label{fig:qualitative_fail}
	\vspace{-0.01in}
\end{figure*}

\parsection{Fully Mask-free Results on OVIS}
Extending from Table 12 of the paper, we further present the results of \modelname on OVIS using COCO box pretraining as initialization in Table~\ref{table:ovis_supp}. Our~\modelname consistently improves the baseline from 10.3 to 13.5 mask AP without using any masks.

\begin{table}[t!]
        \footnotesize
	\centering%
		\caption{Full results of our~\modelname on OVIS~\cite{qi2022occluded} using R50. I: using COCO mask pretrained model as initialization. V: using YTVIS video masks during training.}
		\setlength{\tabcolsep}{2.8pt}
		\vspace{-0.1in}
	\begin{tabular}{lclcccc}
					\toprule
					Method & Mask & AP & AP$_{50}$ & AP$_{75}$ & AR$_{1}$ & AR$_{10}$ \\
					& ann. & & & & & \\
					\midrule
					\underline{\textit{Fully-supervised}:} &  & & &  & \\
					VMT~\cite{vmt} & I+V & 16.9 & 36.4  & 13.7 & 10.4 & 22.7 \\ 
					VITA~\cite{heo2022vita} & I+V & 19.6 & 41.2 & 17.4 & 11.7 & 26.0 \\
					\midrule
					\underline{\textit{Video Mask-free}:} &  & & &  & \\
					VITA~\cite{heo2022vita} + BoxInst~\cite{tian2021boxinst} & I & 12.1 & 28.3 & 10.2 & 8.8 & 17.9 \\
					VITA~\cite{heo2022vita} + \textbf{\modelname} & I & \textbf{15.7}$_{\uparrow\textbf{3.6}}$ & \textbf{35.1} & \textbf{13.1} & \textbf{10.1} & \textbf{20.4} \\
					\midrule
					\underline{\textit{Mask-free}:} &  & & &  & \\
					VITA~\cite{heo2022vita} + BoxInst~\cite{tian2021boxinst} & - & 10.3 & 27.2 & 8.4 & 7.3 & 16.2 \\
					VITA~\cite{heo2022vita} + \textbf{\modelname} & - & \textbf{13.5}$_{\uparrow\textbf{3.2}}$ & \textbf{32.7} & \textbf{10.6} & \textbf{8.8} & \textbf{18.5} \\
					\bottomrule
			\end{tabular}
		\label{table:ovis_supp}
		\vspace{-0.25in}
	\end{table}

\subsection{More analysis on Temporal Correspondence}
\label{sec:supp_vis}

\parsection{Visualization on Temporal Correspondence}
We visualize the dense temporal correspondence matching for TK-Loss computation in Figure~\ref{fig:corres_t}. For better visualization, we randomly sample 100 patch center points from Frame $t$, and plots their respective patch correspondences in Frame $t$+1 using the same color. We observe robust one-to-$K$ patch matching results, especially for the regions near the left leg of the man (inside the red box) and the white frisbee.

\parsection{Correspondence Accuracy}
To further analyze the accuracy rate for the temporal correspondence, since there is no matching ground truth, we adopt the instance masks labels as an approximate measure. 
We randomly take 10\% of the videos from the YTVIS 2019 train set, and split them to 5-frame tube. Following the cyclic connection manner, we compute whether two matched patch center points belonging to the same instance mask label.
The average matching accuracy per image pair is 95.7\%, where we observe the wrong matches are mainly due to the overlapping objects with similar local patch patterns. 

\subsection{More Qualitative Comparisons}
\label{sec:supp_quali}

In Figure~\ref{fig:qualitative_supp}, we provide more qualitative results comparison among Baseline (using spatial losses of BoxInst~\cite{tian2021boxinst}), Ours (using the proposed TK-Loss), and Oracle Mask2Former (trained with GT video and image masks).
Compared to the Baseline, the predicted masks by our approach is more temporally coherent and
accurate, even outperforming the oracle results in some cases (such as the first row of Figure~\ref{fig:qualitative_supp}).
We also identify one typical \textbf{failure case} of our \modelname in Figure~\ref{fig:qualitative_fail}, where the neighboring hand watch and shelf are in almost the same black color, and continuously closing to each other with no sufficient motion information for distinction. We observe even the oracle model trained with GT video masks sometimes fail in correctly delineating these two objects (last row of Figure~\ref{fig:qualitative_fail}). Please refer to the attached video file on our project page for more qualitative results of our \modelname.



\subsection{More Implementation Details}
\label{sec:supp_details}
\parsection{Algorithm Pseudocode} We outline the pseudocode for computing \lossname in Algorithm~\ref{alg:tkloss}, where the execution code does not exceed 15 lines. This further demonstrates the simplicity, beauty and lightness of our TK-Loss without any learnable model parameters.

\parsection{More implementation details} Before computing temporal image patch affinities, we first convert the input image from RGB color space to CIE Lab color space for better differentiating color differences.
We set dilation rate to 3 when performing temporal patch searching.
For the loss balance weights in Equation 7 and Equation 8 of the paper, we set $\lambda_\text{pair}$ to 1.0 and $\lambda_\text{temp}$ to 0.1.
We follow the same training setting and schedule of the baseline methods when integrating our TK-Loss with Mask2Former~\cite{cheng2021mask2former}, SeqFormer~\cite{wu2021seqformer}, VITA~\cite{heo2022vita} and Unicorn~\cite{unicorn} for video instance segmentation training.
When performing mask-free pre-training on COCO with spatial losses of BoxInst, we keep the training details of the integrated method unchanged.
When integrating with Mask2Former using ResNet-50 and batch size 16, the \modelname training on YTVIS 2019 can be finished in around 4.0 hours with 8 Titan RTX. When jointly training with COCO labels, it needs around 2 days.

\newcommand{\assign}{\leftarrow}
\newcommand{\algcomment}[1]{\hspace{0mm}{\footnotesize\# #1}}
\newcommand{\sep}{\,,\quad}

\renewcommand{\algorithmicrequire}{\textbf{Input:}}
\renewcommand{\algorithmicensure}{\textbf{Output:}}

\begin{algorithm}[t]
	\caption{\lossname.}
	\begin{algorithmic}[1]
		\Require Tube length $T$, mask predictions $M$, frame width $W$, height $H$, radius $R$, patch distance threshold $D$. 
		\Ensure TK-Loss $\mathcal{L}_{\text{temp}}$ \\
		\algcomment{$\mathtt{top}K$ denotes selecting top $K$ patch candidates with the maximum patch similarities computed using L$_2$ distance $\mathtt{Dis}(\cdot,\cdot)$}\\
		\algcomment{$L_\mathtt{cons}$ denotes mask consistency loss (Equation 3 of the paper)}
		\State $\mathcal{L}_{\text{temp}} \assign 0.$
		\For{$t = 1, \ldots, T$}
		    \State $\hat{t} \assign (t + 1)\hspace{0.25mm} \% \hspace{0.25mm} T$ \hspace{1.7mm}
		    \State $\mathcal{L}_f^{t\to \hat{t}} \assign 0.$
			\For{$j = 1, \ldots, H\times W$}
			    \State \algcomment{1) Patch Candidate Extraction:}
			    \State $\mathcal{S}_{p_j}^{t\rightarrow \hat{t}} \assign \{\hat{p}_{i}\}_i$, where $\|p_j - \hat{p}_i\| \leq R$
			    \State \algcomment{2) Temporal KNN-Matching:}
			    \State $\mathcal{S}_{p_j}^{t\rightarrow \hat{t}} \assign \mathtt{top}K(\mathcal{S}_{p_j}^{t\rightarrow \hat{t}})$, where $\mathtt{Dis}(p_j, \hat{p}_i) \leq D$
			    \State \algcomment{3) Consistency Loss}
			    \State $\mathcal{L}_f^{t\to \hat{t}} \assign \mathcal{L}_f^{t\to \hat{t}} + \sum_{\hat{p}_i \in \mathcal{S}_{p_j}^{t\rightarrow \hat{t}}} L_\mathtt{cons}(M_{p_j}^t, M_{\hat{p}_i}^{\hat{t}})$
			\EndFor
			\State \algcomment{4) Cyclic Connection}
			\State $\mathcal{L}_{\text{temp}} \assign \mathcal{L}_{\text{temp}} + \mathcal{L}_f^{t\to \hat{t}}/(H\times W)$
		\EndFor
	\end{algorithmic}
	\label{alg:tkloss}
\end{algorithm}


{\small
\bibliographystyle{ieee_fullname}
\bibliography{egbib}
}

\end{document}